\documentclass[journal]{new-aiaa}

\usepackage[utf8]{inputenc}

\usepackage{color}

\usepackage{multirow}
\usepackage{eqparbox}
\usepackage{algorithm}
\usepackage{algpseudocode}
\usepackage{arydshln}
\usepackage{mathtools}
\usepackage{subfigure}
\usepackage{graphicx}
\usepackage{amsmath}
\usepackage[version=4]{mhchem}
\usepackage{longtable,tabularx}
\usepackage{multirow}

\graphicspath{{figures/}}
\newcommand{\floor}[1]{\left \lfloor #1 \right \rfloor}

\usepackage{multirow}

\newcommand{\comments}[1]{}

\title{
	Scan-Matching based Particle Filtering approach for LIDAR-only Localization.
}
\author{Nagavenkat Adurthi \footnote{Assistant Professor, Mechanical and Aerospace Engineering} }
\affil{University of Alabama in Huntsville, Alabama, AL, 35824}

\graphicspath{{figures/}}
\newcommand{\NN}[3]{{\bf \mathcal{N}}(#1:#2,#3)}
\newcommand{\Dim}{\mathbb D}

\newcommand{\Fig}[1]{Figure~\ref{fig1}}

\begin{document}

	\maketitle
	\thispagestyle{empty}
	\pagestyle{empty}

	\begin{abstract}
		This paper deals with the development of a localization methodology for autonomous vehicles using only a $3\Dim$ LIDAR sensor. In the context of this paper, localizing a vehicle in a known 3D global map of the environment is essentially to find its global $3\Dim$ pose (position and orientation) within this map. The problem of tracking is then to use sequential LIDAR scan measurement to also estimate other states such as velocity and angular rates, in addition to the global pose of the vehicle. Particle filters are often used in localization and tracking, as in applications of simultaneously localization and mapping. But particle filters become computationally prohibitive with the increase in particles, often required to localize in a large $3\Dim$ map. Further, computing the likelihood of a LIDAR scan for each particle is in itself a computationally expensive task, thus limiting the number of particles that can be used for real time performance. To this end, we propose a hybrid approach that combines the advantages of a particle filter with a global-local scan matching method to better inform the re-sampling stage of the particle filter. Further, we  propose to use a pre-computed likelihood grid to speedup the computation of LIDAR scans. Finally, we develop the complete algorithm to extensively leverage parallel processing to achieve near sufficient real-time performance on publicly available KITTI datasets.

	\end{abstract}


	\section{Introduction}
	Real-time localization and tracking using sensors such as LIDAR and vision  are becoming essential elements to achieve autonomy for vehicles. While GPS sensors provide a means for accurate localization and tracking, they often produce erroneous measurements in dense urban areas and often no  meaurements altogether in areas such as underpasses, tunnels and parking lots. In the absence of GPS or GPS-denied regions, the problem of localization becomes challenging. To this end, vision and LIDAR sensors provide a means for both localization and tracking. This paper primarily deals with the the problem of localization when only a LIDAR sensor is available. By addressing this base case, it is anticipated that additional sensors such as cameras,IMUs and GPS can easily be integrated and further increase the accuracy and reduce computational complexity. For the problem of localization, it is assumed that a $3\Dim$ (3-dimensional) map of the environment is available. While this assumption of $3\Dim$ maps is demanding,  we believe that with the growth in number of autonomous cars with LIDAR sensors, these maps can be locally built in an offline manner.
	
	\subsection{Background}
	The problem of tracking corresponds to estimating the state of the vehicle from observational data of different sensing modalities. Often, at the core of this tracking process is a filtering algorithm that fuses measurement data with motion models of vehicles. While for linear systems with Gaussian uncertainty, the Kalman Filter \cite{Kalman} is the exact and optimal filter, for nonlinear systems various approximations lead to different filters that trade-off computational complexity with accuracy. Conventional nonlinear filtering algorithms include Extended Kalman Filter (EKF) \cite{moore}, Unscented Kalman Filter (UKF) \cite{jul1}, Gassian Quadrature Filters (GQF) \cite{gfnfp}, Conjugate Unscented Transform Filter \cite{CUTConj_2015} and Particle Filters (PF) \cite{pftut}. Particle filters, and its variants, are often the popular choice for problems of localization, tracking, and even Simultaneous Localization and Mapping (SLAM) \cite{levinson2007map,thrun2002probabilistic}. The ease of implementation of PF and its ability to accommodate nonlinear models makes PF an apt choice. Alternative to the PF, many SLAM problems often use graph based pose optimization approaches to localize the vehicle's position \cite{kummerle2011g,ila2017slam++}    
	
	An intuitive approach for LIDAR-only vehicle localization is to match the current LIDAR scan to an available map. Often the iterative closest point algorithm is used to match the scan or features of the scan to a pre-built map \cite{yoneda2014lidar,moosmann2011velodyne,dube2017segmatch}. Feature based matching can have lower computational complexity but   requires the design of distinguishable features that can uniquely match the scan. A hybrid approach that uses the full grid and  leverages the strengths of feature based approaches is proposed in \cite{choi2014hybrid}, where a Rao-Blackwellized particle filters (RBPF) has been used to utilize the location features.

	Particle filters (PF) have been extensively used for vehicle localization \cite{montemerlo2003fastslam,blanco2019benchmarking,Luca2007,ZHANG2019181}. The conventional PF algorithms are often modified to improve performance. For example, visual lane-marking and GPS are used to update weights of the particles in the PF used for localization \cite{rabe2017robust}. To this end, the primary contributions of this paper involve a novel improvement to conventional particles filters on two fronts: First,  the computations of LIDAR scan likelihoods is made more efficient using sparse pre-computed lookup tables. Second, the re-sampling stage of the particle filter is improved with a scan-to-map matching algorithm. This matching algorithm significantly increases the convergence of the particles to the true vehicle pose. In addition, any loss of tracking is also subsequently corrected during the tracking phase. An algorithmic implementation is described, that also leverages parallel processing to run the scan-to-map matching algorithm along with the particle filter. 
	
	The rest of the paper is organized as follows: In section \ref{pfsec}, the conventional particle filtering approach using LIDAR sensor is revised,  following the initial approach in \cite{blanco2019benchmarking}. In section \ref{mrgl}, the matching algorithm is developed in perspective of localization and tracking. The complete approach is described in section \ref{mainaprch}. In section \ref{sims}, this approach is applied to the KITTI odometry dataset which has the true ground poses for validation. While we assume the mounted LIDAR sensor is a mechanically rotating scanning LIDAR with 360 degrees field of view (FOV) (such as the Velodyne or Ouster), the proposed methodology can also be used for front facing solid state LIDARs with a smaller FOV, and stereo camera based pseudo-LIDAR point clouds \cite{wang2019pseudo}.
	
	\subsection{Notations}
	A vector is denoted using a bold alphabet such as $\mathbf{x}_k$, where the subscript $k$ indicates the discrete time instant $t_k$. $p(\mathbf{x}_k)$ is used to represent a probability density function (PDF) and $\NN{\mathbf{x}}{\mu}{P}$ represent a multivariate Gaussian PDF for the random vector $\mathbf{x}$ with mean $\mu$ and covariance matrix $P$. The superscript in $\mathbf{x}^{(i)}_k$ is used to denote the $i^{th}$ particle/sample of the particle filter. The $3\Dim$ position vector of the vehicle is represented as $\mathbf{p}=[x,y,z]^T$ and the orientation of the vehicle is described by the yaw-pitch-roll angle vector $\boldsymbol{\zeta}=[\theta,\phi,\psi]^T$. The pose of the vehicle is represented by the vector $\boldsymbol{\xi}=[x,y,z,\theta,\phi,\psi]^T=[\mathbf{p}^T,\boldsymbol{\zeta}^T]^T$ and the corresponding  homogeneous transformation matrix is represented as 
	\begin{align}
	H(\boldsymbol{\xi})\equiv\begin{bmatrix}
	R(\boldsymbol{\zeta})&\mathbf{p}\\\mathbf{0}_{1\times 3}&1
	\end{bmatrix}
	\end{align}
	where $R(\boldsymbol{\zeta})$ is the rotation matrix for the z-y-x (yaw-pitch-roll) Euler angles. It is assumed that the body fixed x-axis is pointing in the forward direction of the vehicle and the corresponding body-fixed z-axis is pointing vertically up. A $3\Dim$ LIDAR scan at time $t_k$ is denoted as $\mathbf{X}_k$ and a single point with index $j$ in this scan is represented as ${X}^{(j)}_k$. The global $3\Dim$ map is represented as $\mathcal{M}$, the corresponding $2\Dim$ top-view or bird's eye view (BEV) of the global map is represented as $\mathcal{M}_{bev}$.  A point in a BEV LIDAR scan will be denoted as $B^{(j)}_{k}$, which is computed by setting the z-coordinate (assuming z-axis points vertically up) of ${X}^{(j)}_k$ to zero.
	 
	\section{Particle filtering}\label{pfsec}
	Consider the vehicle dynamical model and mounted sensor model as:
	\begin{align}
	\mathbf{x}_{k}=f(\mathbf{x}_{k-1})+\omega_{k-1}\\
	\mathbf{y}_{k}=h(\mathbf{x}_{k})+\nu_{k}
	\end{align}
	where $\mathbf{x}_{k}$ is the state vector  at time step k and  $\mathbf{y}_{k}$ is the measurement vector at time step $k$. $\omega_{k-1}$ is the process noise and $\nu_{k}$ is the measurement noise, both modeled as Gaussian white noise sequences with covariances $Q_{k-1}$ and $R_{k}$, respectively.   In general, this measurement vector $\mathbf{y}_{k}$ can directly hold the raw LIDAR scans, IMU measurements, or features extracted from LIDAR scans and images. The general filtering algorithm works by solving the two step filter equations, namely the Chapman-Kolmogorov-Equation (CKE) and the Bayes' Rule \cite{jazwin}:
	\begin{align}
	\textit{Propagation (CKE):}&\quad p(\mathbf{x}_{k}|Y_{(1:k-1)})=\int p(\mathbf{x}_{k}|\mathbf{x}_{k-1})p(\mathbf{x}_{k-1}|Y_{(1:k-1)}) \:d\mathbf{x}_{k-1} \label{CKE}\\
	\textit{Measurement Update (BR) :}&\quad p(\mathbf{x}_{k}|Y_{(1:k)}) = \frac{p(\mathbf{y}_{k}|\mathbf{x}_{k},Y_{(1:k-1)}) p(\mathbf{x}_{k}|Y_{(1:k-1)})}{p(Y_{(1:k)})} \label{BR}
	\end{align}
	where $p(\mathbf{x}_{k}|Y_{(1:k-1)})$ is the \emph{a priori} state probability density function (PDF), $p(\mathbf{x}_{k}|Y_{(1:k)})$ is the \emph{a posteriori} PDF after the measurement update  and $Y_{(1:k-1)}$ is the sequence or history of measurements until time step $k-1$.
	The particle filter \cite{pftut} solves these equations by approximating the state PDF using finite set of samples or particles as $p(\mathbf{x}_{k}|Y_{(1:k)})\approx \sum_{i}^{N_s}w_{k}^{(i)}\delta(\mathbf{x}-\mathbf{x}_{k}^{(i)})$. Hence, the PF only keeps track of the set of samples and their corresponding weights $\{\mathbf{x}_k^{(i)},w^{(i)}_k\}_{i=1,N_p}$ at every time step $k$. Propagation of the particles by \eqref{CKE} reduces to sampling from the transition PDF \cite{pftut}, 
	\begin{align}
	\mathbf{x}^{(i)}_{k}\sim p(\mathbf{x}_{k}|\mathbf{x}^{(i)}_{k-1})\equiv \NN{\mathbf{x}_{k}}{f(\mathbf{x}^{(i)}_{k-1})}{Q_{k-1}}
	\end{align}
	where $\NN{\mathbf{x}}{\mu}{P}$ represents a multivariate Gaussian PDF for the random variable $\mathbf{x}$ with mean $\mu$ and covariance matrix $P$.  Here the proposal density is taken as $p(\mathbf{x}_{k}|\mathbf{x}^{(i)}_{k-1})$, though not optimal it is often the most pragmatic choice \cite{pftut}. The weights remain the same during propagation. The \emph{a posteriori} PDF is approximated by updating the weights of the samples using the measurement likelihood model of the sensors as \cite{pftut}:
	\begin{align}
	w^{(i)}_k \propto  p(\mathbf{y}_{k}|\mathbf{x}^{(i)}_{k}) w^{(i)}_{k-1}
	\end{align} 
	Hence, the resultant set is denoted as $\{\mathbf{x}_k^{(i)},w^{(i)}_k\}_{i=1,N_p}$. The measurement likelihood is typically taken as $p(\mathbf{y}_{k}|\mathbf{x}^{(i)}_{k}) =\NN{\mathbf{y}_{k}}{h(\mathbf{x}^{(i)}_{k})}{R_k}$ when the sensor model $h(\mathbf{x}^{(i)}_{k})$ is well-defined and simple to evaluate. Though simple, the practical implementation of PF requires the proper tuning of parameters such as the number of particles and the proposal density. Further, PFs are susceptible to the problem of sample degeneracy,  where after a few iterations only a small fraction of the samples have significant weight. This problem is solved by using better proposal densities or resampling strategies \cite{pftut}.  The conventional resampling procedure is that of bootstrap filtering. Ideally, increasing the number of particles can reduce the effects of degeneracy and sample impoverishment, but it directly leads to large computational complexity. While other resampling procedures such as regularized particle filters can also be used, they are computationally demanding \cite{pftut}.  Hence, it is imperative that we  reduce the number of particles and improve resampling strategies to achieve better performance of particle filters. 
	\subsection{Dynamic model}\label{pfsecdynmodel}
	The dynamical model used for tracking the vehicle is modeled as a constant acceleration and constant turn-rate model. The state vector for the vehicle is $\mathbf{x}=[x,y,z$ $,\theta,\phi,\psi$ $,v,\dot{\theta},\dot{\phi},\dot{\psi},a]^T$, where $\mathbf{p}=[x,y,z]^T$ is the position (with units in meters) of the vehicle in the reference frame of the map, $\boldsymbol{\zeta}=[\theta,\phi,\psi]^T$ are the yaw-pitch-roll angles (measured in radians) of the body fixed frame of the vehicle with respect to the map reference frame. The body fixed frame is centered at the origin of the LIDAR sensor with the x-axis pointing in the forward direction of the vehicle. $v$ is the magnitude of the velocity of the vehicle along the body fixed x-axis. $\boldsymbol{\dot{\zeta}}=[\dot{\theta},\dot{\phi},\dot{\psi}]^T$ are the corresponding yaw-pitch-roll rates and $a$ is the magnitude of acceleration of the vehicle. Given the state $\mathbf{x}_{k-1}=[\mathbf{p}^T_{k-1},\boldsymbol{\bar{\zeta}}^T_{k},v_{k-1},\boldsymbol{\dot{\zeta}}^T_{k-1},a_{k-1}]^T$ at time step $k-1$, the discrete-time dynamical model to propagate the state vector to time step $k$, denoted as $\mathbf{x}_k=[\mathbf{p}^T_{k},\boldsymbol{\zeta}^T_{k},v_k,\boldsymbol{\dot{\zeta}}^T_{k-1},a_{k-1}]^T$ is given by the approximate first order hold model:  
	\begin{align}
	R_{k-1}&=R(\theta_{k-1},\phi_{k-1},\psi_{k-1}) \label{dyneq1}\\
	\hat{v}_{k-1} &= R_{k-1}[1,0,0]^T\\
	\mathbf{p}_{k} &= \mathbf{p}_{k-1}+v_{k-1}\hat{v}_{k-1}\:\Delta t +\frac{1}{2}a_{k-1}\hat{v}_{k-1}\:\Delta t^2 \\
	v_k&=v_{k-1}+a_{k-1}\Delta t\\
	\boldsymbol{\zeta}_{k}&=\boldsymbol{\zeta}_{k-1}+\boldsymbol{\dot{\zeta}}_{k-1}\Delta t \label{dyneqN}
	\end{align}
	here $R(\theta_{k-1},\phi_{k-1},\psi_{k-1})$ is the rotation matrix computed using the yaw-pitch-roll angles. In this model, the angular rates and the acceleration are assumed constant for short discrete time-steps.
	
	\subsection{Sensor model}\label{pfsecSensmodel}
	The next problem is that of evaluating the the likelihood probability $p(\mathbf{y}_{k}|\mathbf{x}^{(i)}_{k})$ for LIDAR scans in a given map. We follow the model in \cite{blanco2019benchmarking} to compute the likelihood. For a sample vehicle state $\mathbf{x}_k^{(i)}$, we use the notation $R^{(i)}_k=R(\theta^{(i)}_k,\phi^{(i)}_k,\psi^{(i)}_k)$ and $\mathbf{p}^{(i)}_k=[x^{(i)}_k,y^{(i)}_k,z^{(i)}_k]$ for the corresponding  the rotation matrix and translation vector, respectively for the vehicle pose. The LIDAR scan at time step $k$ is denoted as $\mathbf{X}_{k}=[\ldots, X^{(j)}_k,\ldots]^T$ for $j=1,2,\ldots,N_s$, where $N_s$ is the number of points in the scan. The scan likelihood model (sensor model) is  constructed as:
	\begin{align}
	\bar{X}^{(j)}_k&=R^{(i)}_k X^{(j)}_k+\mathbf{p}^{(i)}_k\\
	d_j&\leftarrow \textit{NNdist}(\bar{X}^{(j)}_k,\mathcal{M}) \label{nndist}\\
	p(\mathbf{y}_{k}|\mathbf{x}^{(i)}_{k})&\approx exp(-\sum_{j=1}^{N_s}\frac{\min(d_j^2,d^2_{max})}{\sigma^2})
	\end{align}
	In other words, the pose of the particle $\mathbf{x}^{(i)}_{k}$ is used to transform the LIDAR scan points from the sensor reference frame to the world inertial frame and then their corresponding distance to their closest neighbors in the map are determined. The likelihood is simply constructed as a measure of how close the transformed LIDAR scan is to the Map. This likelihood function is strongly influenced by the chosen parameters of maximum distance $d_{max}$ and standard deviation $\sigma$. $NNdist(x,\mathcal{M})$ is the distance between a point $x$ and its closet neighbor in the map $\mathcal{M}$. The nearest distance can be efficiently computed by kdtrees and octrees. But it is to be noted that a typical LIDAR scan contains more than $N_s~100K$ points. When combined with large number of particles ($N_p~1000-10,000$),  the number of nearest neighbor searches ($\sim N_sN_p$) becomes computationally intractable. To this end, authors in \cite{blanco2019benchmarking} propose to decimate the number of points in the LIDAR scan to achieve better real-time performance. In other words, the number of points in the LIDAR scan is reduced by taking  every $n^{th}$ point from the set of $N_s$ points in the scan. The parameter $n$ is chosen by the user to reduce computational complexity. The bench-marking results in \cite{blanco2019benchmarking} show sufficient localization capability with a much smaller subset of points of the LIDAR scan, but with large number of particles. To this end, we use this benchmarks as a guide and instead use  voxel-based down-sampling to reduce the number of points uniformly over the $3\Dim$ space. In voxel-based down-sampling, only one sample is randomly chosen within each voxel. This allows the LIDAR scan to be uniformly distributed over the original  LIDAR scan.
	
	But as the number of particles increase, the computational complexity remains significantly high. It can be observed that the nearest neighbor search routine $NNdist$ in \eqref{nndist} is the main computational bottleneck, particularly when the LIDAR scan has thousands of points. To ease this computational complexity,  a pre-computed sparse $3\Dim$ tensor is used to store the nearest distance to a map $\mathcal{M}$. This way, the pre-computed nearest neighbor distance to the map can be used for all LIDAR scans. The $3\Dim$ grid is determined by estimating the lower bound point  $\bar{lb}=[lb_x,lb_y,lb_z]^T=min(\mathcal{M})$ and upper bound point $\bar{ub}=[ub_x,ub_y,ub_z]^T=max(\mathcal{M})$ for the bounding box of the map $\mathcal{M}$, with length along the axes  denoted as $\bar{D}=[D_x,D_y,D_z]^T=\bar{ub}-\bar{lb}$. Let the desired resolution of this spatial grid be denoted as $\bar{\delta}=[\delta_x,\delta_y,\delta_z]$. Assuming a uniform grid of points on each axis, the $3\Dim$ points are simply constructed as a tensor product 
	\begin{figure}
		\begin{center}
			\includegraphics[trim=8cm 0cm 8cm 0cm,clip=true,width=3in]{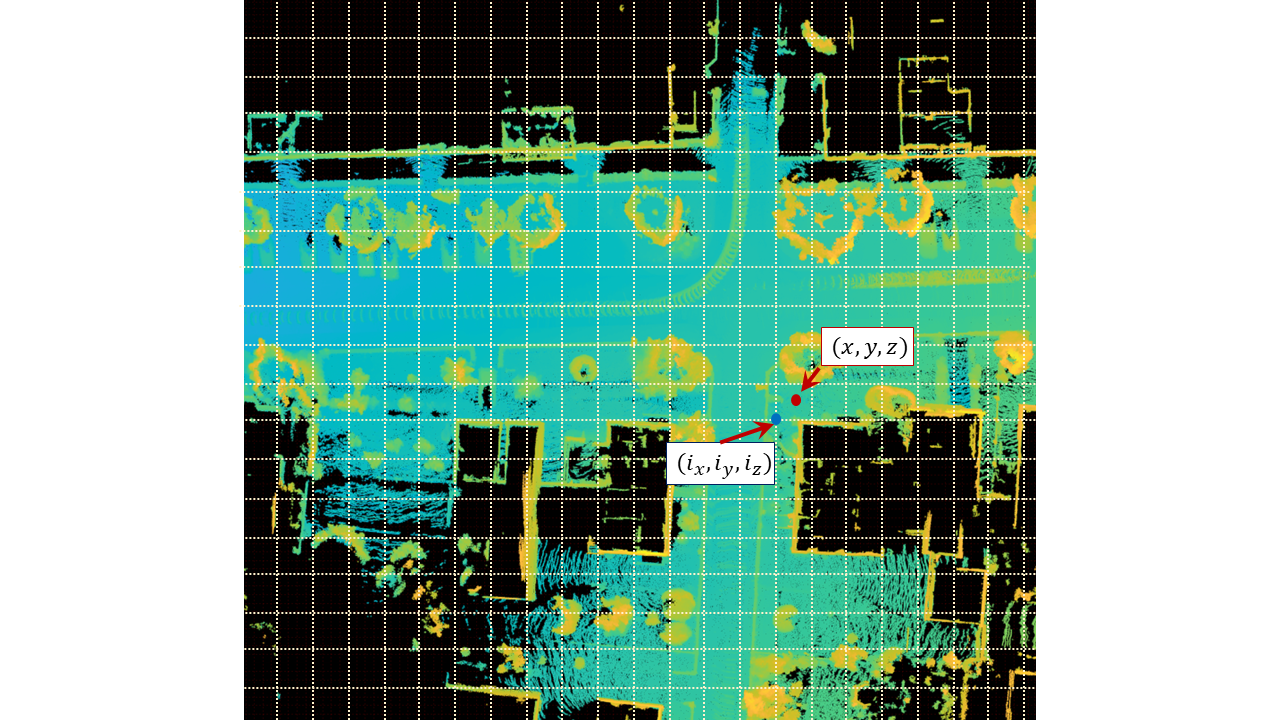}
		\end{center}
		\caption{Grid of nodes over a crossroad section of the map (top view). Dark regions have no map points.}\label{lookupNNdist2}
	\end{figure}
	as $\mathcal{G}=\left\{\{x(i_x) \}\otimes\{y(i_y) \}\otimes \{z(i_z) \}\right\}$, where the grid nodes along the x-axis are represented as $x(i_x)$ $=$ $lb_x+i_x\delta_x$ with $i_x=0,1,\ldots,\floor{\frac{D_x}{\delta_x}}-1$. $\{y(i_y) \}$ and $\{z(i_z) \}$ are similarly defined for the $y$ and $z$ dimensions respectively.  
	
	Figure \ref{lookupNNdist2} shows an instance of this uniform grid nodes over a section of the map (top view). Here, the map $\mathcal{M}$ is constructed by fusing sequential LIDAR scans. The nearest point in the map $\mathcal{M}$ to a grid point $(i_x,i_y,i_z)$  is pre-computed along with its distance. Now, given any LIDAR scan point $(x,y,z)$, as shown in Figure \ref{lookupNNdist2}, it can be decimated to its corresponding grid node $(i_x,i_y,i_z)$. This way, multiple points that fall within the cell located between corners $(i_x,i_y,i_z)$ and $(i_x+\delta_x,i_y+\delta_y,i_z+\delta_z)$ can be mapped to the same grid node and thus save on the number of nearest neighborhood searches. The underlying assumption is that the nearest distance of points in a cell to the map can be approximated by the nearest distance of its corner $(i_x,i_y,i_z)$ to the map. The error in the nearest neighborhood distance is a function of the grid resolution $\bar{\delta}$ used to create the cells.
	
	As a regular grid is used, not all the grid nodes are necessary as the map is usually less dense in regions where the vehicle cannot navigate (for example the dark regions in Figure \ref{lookupNNdist2}). One approach is to use an adaptive grid, where higher resolution grids are only used in dense regions of the map. Alternatively, in this work, we use a high resolution regular grid but instead use a sparse lookup table, denoted as $\mathcal{L}$, to store the nearest neighbors to the map. The sparsity of map is further increased by only saving grid points that have a neighbor in the map within a distance of $d_{max}$. The sparse map $\mathcal{L}$ is populated offline and offers fast retrieval of approximate nearest neighborhood distances. In the offline process to construct the lookup table, the distance $d$ of each grid node index $(i_x,i_y,i_z)$, with $0\le i_x<\floor{\frac{D_x}{\delta_x}}$, $0\le i_y<\floor{\frac{D_y}{\delta_y}}$ and $0\le i_z<\floor{\frac{D_z}{\delta_z}}$ is computed using the nearest neighborhood search as $d=\textit{NNdist}([x(i_x),y(i_y),z(i_z)]^T,\mathcal{M}) $. Then this value is inserted into to the sparse map as $\mathcal{L}(i_x,i_y,i_z) \leftarrow d$ when  $d<d_{max}$. Now given any $3\Dim$ point $(x,y,z)$, the retrieval of its approximate closest distance to the map is achieved by first computing its corresponding grid node index as:
	\begin{align}
	 i_x = \floor{\frac{x-lb_x}{\delta_x}}, \quad 	 i_y = \floor{\frac{y-lb_y}{\delta_y}}, \quad  i_z = \floor{\frac{z-lb_z}{\delta_z}}
	\end{align}
	If the index triplet $(i_x,i_y,i_z)$ exists in $\mathcal{L}$, it can be retrieved as $d=\mathcal{L}(i_x,i_y,i_z)$. If $(i_x,i_y,i_z)$ does not exist, then the default $d_{max}$ is directly used. It can be observed that the closest distance retrieval is fast as it involves only simple lookup of keys. In this work we use the parallel sparse hash map in \cite{parahash}. With this fast lookup, the likelihood of  LIDAR scans with large number of points can be quickly computed. In the next section, a scan matching method to improve the particle filter is discussed.
	
	\section{Multi-Resolution Scan Matching}\label{mrgl}
The multi-resolution scan matching method used in this work is adopted from \cite{olson2015m3rsm}. Rather than using it for scan-to-scan matching,  the approach is used for scan-to-map matching. Only a brief review of the approach is discussed here and complete details can be found in \cite{olson2015m3rsm}.  A similar approach of branch bound method for scan matching and identifying loop closures can be found in \cite{hess2016real} . Henceforth, the scan and map will refer to a point-cloud in $2\Dim$, typically the top-view or the bird's eye view of the $3\Dim$ scan. This $2\Dim$ bird's eye view map will be referred to as $\mathcal{M}_{bev}$.

The points in a LIDAR scan are often given in the LIDAR-sensor fixed frame or even the body fixed. The objective is then to find the homogeneous transformation $T$ that takes these LIDAR scan points and aligns them to the map in the map-fixed reference frame. As the points are in $2\Dim$, it is only required to compute the $2\Dim$ rotation matrix $R_z(\theta)$ and the translation $\mathbf{t}_{(x,y)}$. Conventional algorithms such as iterative closest point ICP and GICP \cite{segal2009generalized} are predominantly local in nature, and usually work well in the immediate neighborhood of the initial guess. The cost of aligning the scan to the map, as in the iterative closest point, is defined as $\min\limits_{\theta,\mathbf{t}_{x,y}}\sum_{j=1}\left\{m^{(j)}-\left(R_z(\theta)B^{(j)}+\mathbf{t}_{(x,y)}\right) \right\}^2$ where $B^{(j)}$ is a point in the scan ($2D$ BEV projection) and $m^{(j)}\in\mathcal{M}_{bev}$ is the closest point to the transformed point $R_z(\theta)B^{(j)}+\mathbf{t}_{(x,y)}$. This cost function is nonlinear and often results in local minima,  that tend to be wrong matches. Alternatively, the only way is to use a brute force approach or an exhaustive search approach by enumerating all possible angles $\theta$ and translations $\mathbf{t}_{(x,y)}$ to find the global minimum. But this is computationally challenging and often intractable. To this end, the method in \cite{olson2015m3rsm} uses a carefully designed exhaustive search method, where the infeasible or high cost regions of search space are progressively eliminated using low resolution grids of the search space, thus reducing computational complexity. 

\begin{figure}
		\begin{center}
			\includegraphics[trim=0cm 1cm 1cm 0cm,clip=true,width=5in]{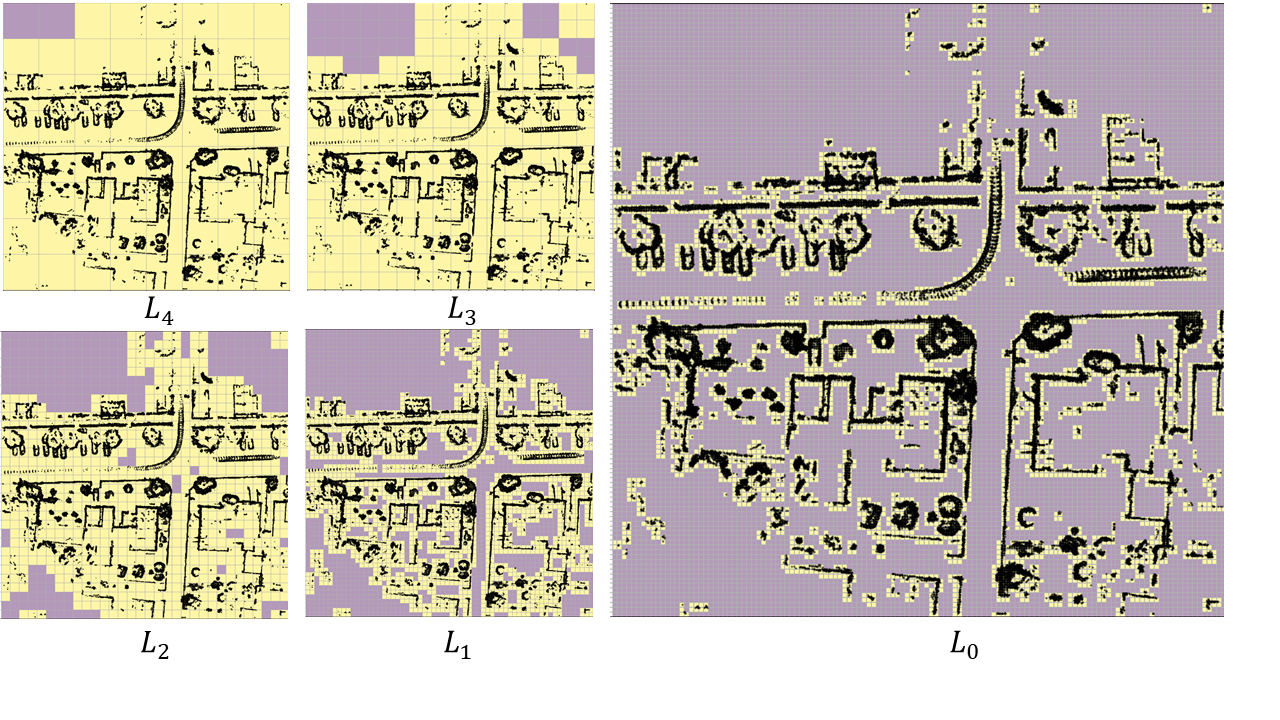}
		\end{center}
		\caption{Examples of levels ($L_0$-$L_4$) for a section of the BEV map }\label{scan2maplevelseg}
	\end{figure}

Starting with highest desired match resolution $\mathbf{d}_{m}=[dx_{m},dy_{m}]^T$, a $2\Dim$ regular grid is constructed for the map $\mathcal{M}_{bev}$ with spacing $\mathbf{d}_{m}$ and denoted as level $L_0$. A scan is assumed to be matched to the map, if  all the scan points lie within a distance of $\mathbf{d}_{m}$ from the map. The value within a cell with index $\mathbf{q}$, denoted as $L_0(\mathbf{q})$, with lower bound $\mathbf{x}_\mathbf{q}=[x_\mathbf{q},y_\mathbf{q}]^T$ and upper bound $\mathbf{x}_\mathbf{q}+\mathbf{d}_{m}$ is taken as the number of map points that fall within this cell.  A simpler approach would be to just use a binary value to indicate if a cell contains at least one map point. Progressive levels $L_l$ for $l=1,2,\ldots$ are then constructed by using $2\times 2$ max-pooling operation with a stride of two, forming a sort of image pyramid with corresponding grid spacing denoted as $\mathbf{d}_l$ at level $l$.  Figure \ref{scan2maplevelseg}  shows an example of these progressive levels for a desired match resolution of $\mathbf{d}_{m}=[2m,2m]^T$ at $L_0$. Subsequent levels $L_1$, $L_2$, $L_3$ and $L_4$ have double the resolution at each level as seen in  Figure \ref{scan2maplevelseg}. The lighter shade grid cells (yellow in Figure \ref{scan2maplevelseg}) correspond to cells with at least one point and the darker shade cells have no map points within them and hence have a value of zero.
The matching score for a given transformation $\{\theta,\mathbf{t}_{(x,y)}\}$ at a particular level $l$, is given as 
\begin{align}
\max\limits_{\theta,\mathbf{t}_{(x,y)}} J_l(\theta,\mathbf{t}_{(x,y)}) =\sum_{j=1}{L}_l\left(R_z(\theta)B^{(j)}+\mathbf{t}_{(x,y)}\right) \label{s2mcost11}
\end{align}

Let $\mathbf{q}$ be the index of the cell which contains the point $R_z(\theta)B^{(j)}+\mathbf{t}_{(x,y)}$, then ${L}_l$ in \eqref{s2mcost11} is evaluated as $L_l(\mathbf{q})$ and otherwise zero. This cell index $\mathbf{q}$  is directly computed as $\floor{\frac{R_z(\theta_r)B^{(j)}+\mathbf{t}_{(x,y)}}{\mathbf{d}_l}}$. Thus the total score computation in \eqref{s2mcost11}  corresponds to a simple lookup in $L_l$. The objective is then to find the optimal transformation that maximizes this nonlinear score, as the highest score would correspond to the best alignment to the map.  To avoid the nonlinearity due to the angular rotation, one can consider  discrete angular positions  as $\theta_r=-\pi+rd_{\theta}$ for $r=[0,1,\ldots,2\pi/d_{\theta})$, where $d_{\theta}$ is the desired angular resolution. The subsequent optimization problem is then to maximize \eqref{s2mcost11} only with respect to $\mathbf{t}_{x,y}$ as $J^{(r)}_l(\mathbf{t}_{x,y})=\max\limits_{\mathbf{t}_{x,y}} \sum_{j=1}L_l\left(R_z(\theta_r)B^{(j)}+\mathbf{t}_{(x,y)}\right)$ individually for each $\theta_r$. While it can be observed that the cost is piece-wise linear in $\mathbf{t}_{(x,y)}$, local search algorithms that rely on gradients often result in local minima. A global search is generally required and it tends to become computationally demanding for each $\theta_r$. To this end, an intuitive approach for exhaustive search over $\mathbf{t}_{(x,y)}$ is to quickly eliminate cases that do not lead to the optimal solution. This exhaustive search for all $\theta_r$ can be done sequentially beginning at coarse resolutions of $\mathbf{d}_l$ and progressing toward the finer resolutions ($l=0$).     At low resolution levels, the spatial grid is coarse and configurations that do not lead to the optimal one can be quickly pruned at early stages leading to a significant speedup. At each level, it suffices to translate $\mathbf{t}_{(x,y)}$ by the resolution of the grid thus further speeding up the search process. As discussed in \cite{olson2015m3rsm}, a max-heap data structure, ordered using the score of a configuration, is most appropriate to refine the search over levels with higher resolution. In addition to the score, each entry in the heap also saves $\theta_r$, the level $l$, the lower bound and  length of the rectangular search-region. At each subsequent higher resolution level, this rectangular search-region is split into four quadrants and pushed into the heap. When the top of the heap is at level $l=0$ corresponding to $\mathbf{d}_m$, the optimal solution is returned. While this matching process is efficient than an exhaustive search, it is still computationally prohibitive when used to match a scan against the full map. In the next section, the main approach is discussed that leverages this hierarchical scan-to-map matching approach in an efficient manner to improve the localization process of the particle filter (section  \ref{mainaprchmatch}).

	\section{Main approach} \label{mainaprch}
\begin{figure}
		\begin{center}
			\includegraphics[width=5in]{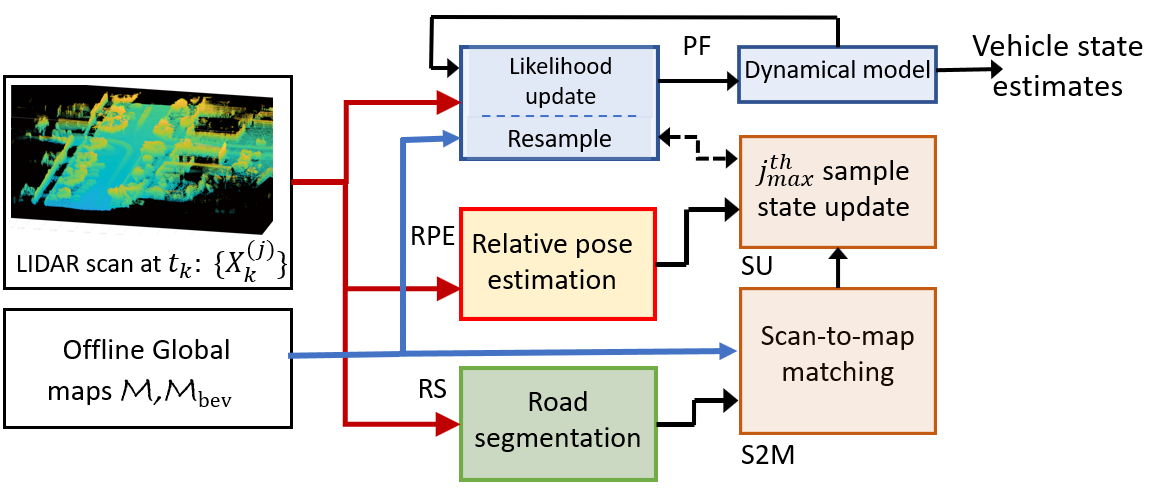}
		\end{center}
	\caption{Block diagram of the proposed approach}\label{flchart}
	\end{figure}

	The overall approach is illustrated in Figure \ref{flchart} for a single time step, where each block is used to designate a specific operation and the arrows connecting the blocks show the flow of information. These main blocks are particle filter block (PF), relative pose estimation block (RPE), road segmentation block (RS), scan-to-map matching block (S2M) and the sample update block (SU). The blocks are implemented in separate threads and asynchronously. Specifically, the PF block running independently on a thread, involves the computation of likelihood probabilities for the samples, propagating the samples and resampling. The PF block is the primary block resulting in the vehicle state estimates.  The convergence and robustness of the PF block is enhanced by the SU and S2M blocks, that enrich the samples/particles with the high weights. Thereby, subsequent resampling will lead to better samples for the vehicle state.  The SU and S2M blocks run on the same thread, updating the sample as soon as the the scan-to-map matching is complete in the S2M block. The RPE block is run on a separate thread and computes the scan-to-scan poses and the corresponding linear and angular velocity. Finally, the RS block works in a background thread to generate LIDAR scans with the road removed. It is necessary to remove the road (and even flat surfaces) in the bird's eye view of the map $\mathcal{M}_{bev}$ and the LIDAR scan as they do not contain features for alignment. The primary features in the top view of the map are often the building, walls and sidewalk edges that strongly aid the scan-to-map matching algorithm in section \ref{mrgl}. Though the S2M block is computationally intensive,  it generates very accurate global pose alignment that outweighs its inherent complexity. The accurate global pose alignment significantly enhances the  PF, making it accurate and even robust. The details of each block are further described below.
	\subsection{Particle filter}
	The particle filter implements the sample propagation using the dynamical model in section \ref{pfsecdynmodel} and updates the sample weights using the likelihood approach as detailed in section \ref{pfsecSensmodel}. The particle filter is based on the sequential importance sampling (SIS) \cite{pftut}  particle filtering algorithm, where a bootstrap resampling procedure is implemented. The resampling step is crucial to address the degeneracy problem of particle filters. The likelihood probability considered in \ref{pfsecSensmodel} is highly dependent on $d_{max}$, $\sigma$ and the pose of the vehicle, particularly the $[x,y]$ and yaw states. In many cases where the vehicle suddenly makes turn or accelerates, all but few samples with pose close to the true vehicle pose will have significant weights thus rendering the rest of the samples useless. Resampling addresses this issue by generating samples close to the samples with significant weight. Further, the process noise matrix $Q_{k}$ in \ref{pfsecdynmodel} plays a crucial role in improving the diversity of the samples by adding randomness to the  states of the particles, and thus helps to capture vehicle turns and rapid change in accelerations (slowing, stopping and speeding). Further, by having adequate process noise, the problem of sample impoverishment \cite{pftut} is reduced, without which the particles will collapse to a single particle after few iterations.  The resampling step is only performed when the effective sample size $N_{eff}$ is less than $50\%$ \cite{pftut}.

	\subsection{Relative pose estimation}\label{mainaprchrelstate}
	The relative pose estimation (RPE) block computes the relative poses between consecutive time steps. Scan-to-scan matching is done in a separate thread to estimate these relative poses. As the current work focuses on LIDAR-only approach, these relative pose estimates provide an alternate way to approximately compute velocity and angular rates that help in the overall tracking process using the particle filter. Ideally, an inertial measurement unit (IMU) sensor can directly provide angular rates and can replace this subsystem to further save computational resources. For $3\Dim$ LIDAR scans, we use the Generalized Iterative Closest Point (GICP) algorithm \cite{segal2009generalized} for its robustness in assessing the full $3\Dim$ relative pose. While there is computational overhead in computing the covariance matrices for all points, the GICP can have better stability and accuracy compared to the conventional Iterative Closest Point/Plane algorithms that are often sensitive to outliers \cite{9336308}. We use voxel-based downsampling to reduce computational complexity and subsequently RANSAC algorithm to further introduce robustness against outliers. The GICP (or other scan-to-scan matching algorithms) computes the relative pose $H_{(k,k-1)}$ that aligns the LIDAR scan at timestep k with the LIDAR scan at timestep $k-1$.
The pose from initial time is then computed as 
	\begin{align}
	H_k = H_{(k,k-1)}H_{(k-1,k-2)}\cdots H_{(1,0)}H_0 \equiv H_{(k,0)}H_0
	\end{align}
	where $H_0$ is the initial estimate of the vehicle's pose at time step $0$. As $H_0$ is unknown and is estimated later on through the main filter, one can assume $H_0$ to be identity. The velocity is computed using simple one step Euler backward difference from the pose estimates as $\mathbf{v}_k \approx (\mathbf{p}_{k}-\mathbf{p}_{k-1})/\Delta t$ and $\boldsymbol{\dot{\zeta}}_k \approx (\boldsymbol{\zeta}_{k}-\boldsymbol{\zeta}_{k-1})/\Delta t$ where $[\mathbf{p}_{k},\boldsymbol{\zeta}_{k}]^T$ and $[\mathbf{p}_{k-1},\boldsymbol{\zeta}_{k-1}]^T$ are the poses extracted from $H_{k}$ and $H_{k-1}$, respectively. Better estimates can be achieved by using a cubic spline interpolation \cite{de1978practical}. Note that, these estimates are noisy on account of simple backward difference for derivative estimates. As such, these estimates can be assumed as pseudo measurements for the particle filter.
	
	\subsection{Road segmentation}\label{mainaprchroadseg}
\begin{figure}
			\centering

				\subfigure[Top view]{\includegraphics[trim=12cm 0cm 11cm 0cm,clip=true,width=2.3in]{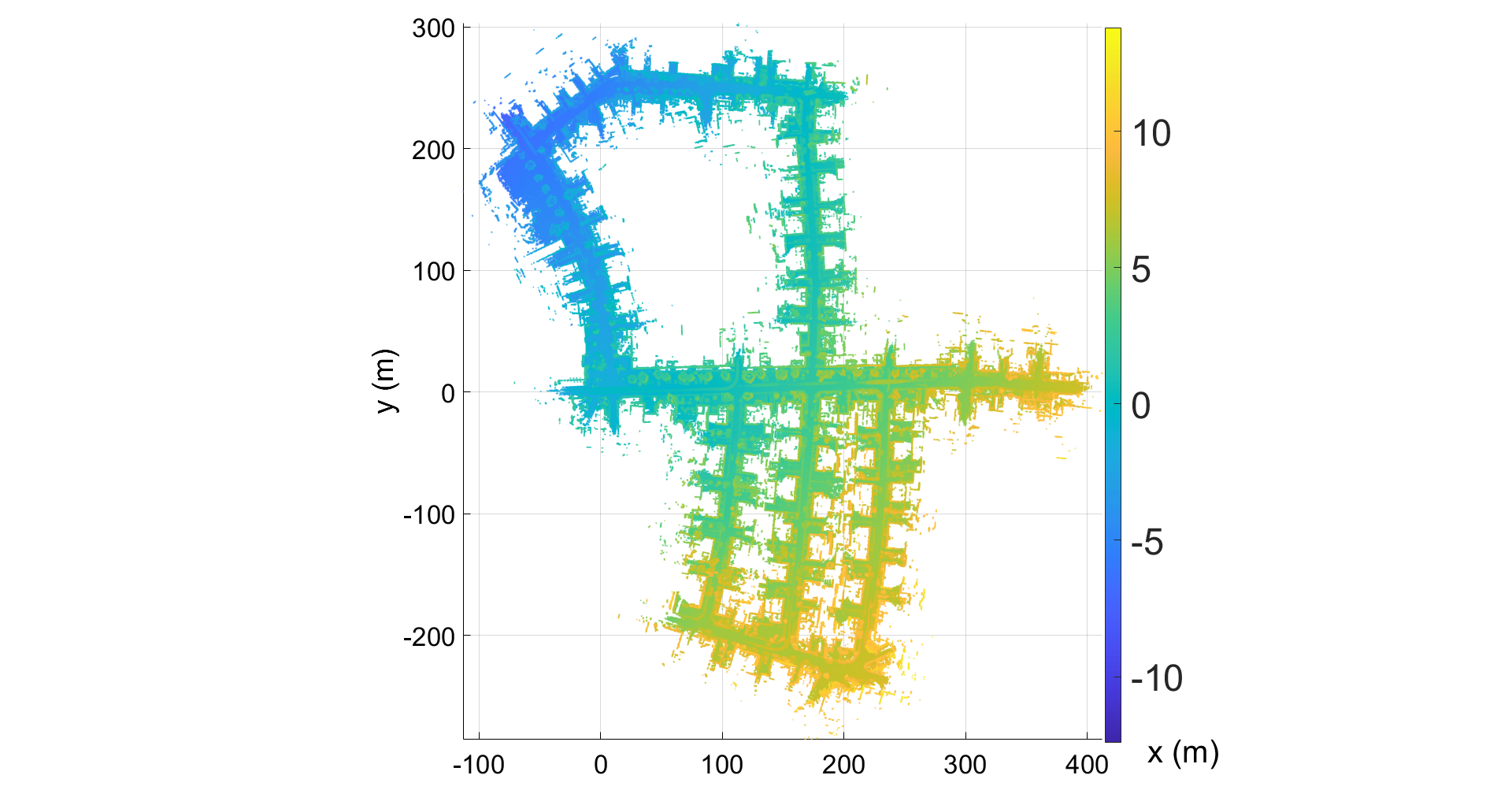}\label{pclroadrem3}}
				
				\subfigure[Top view  (road removed)]{\includegraphics[trim=12cm 0cm 12cm 0cm,clip=true,width=2.5in]{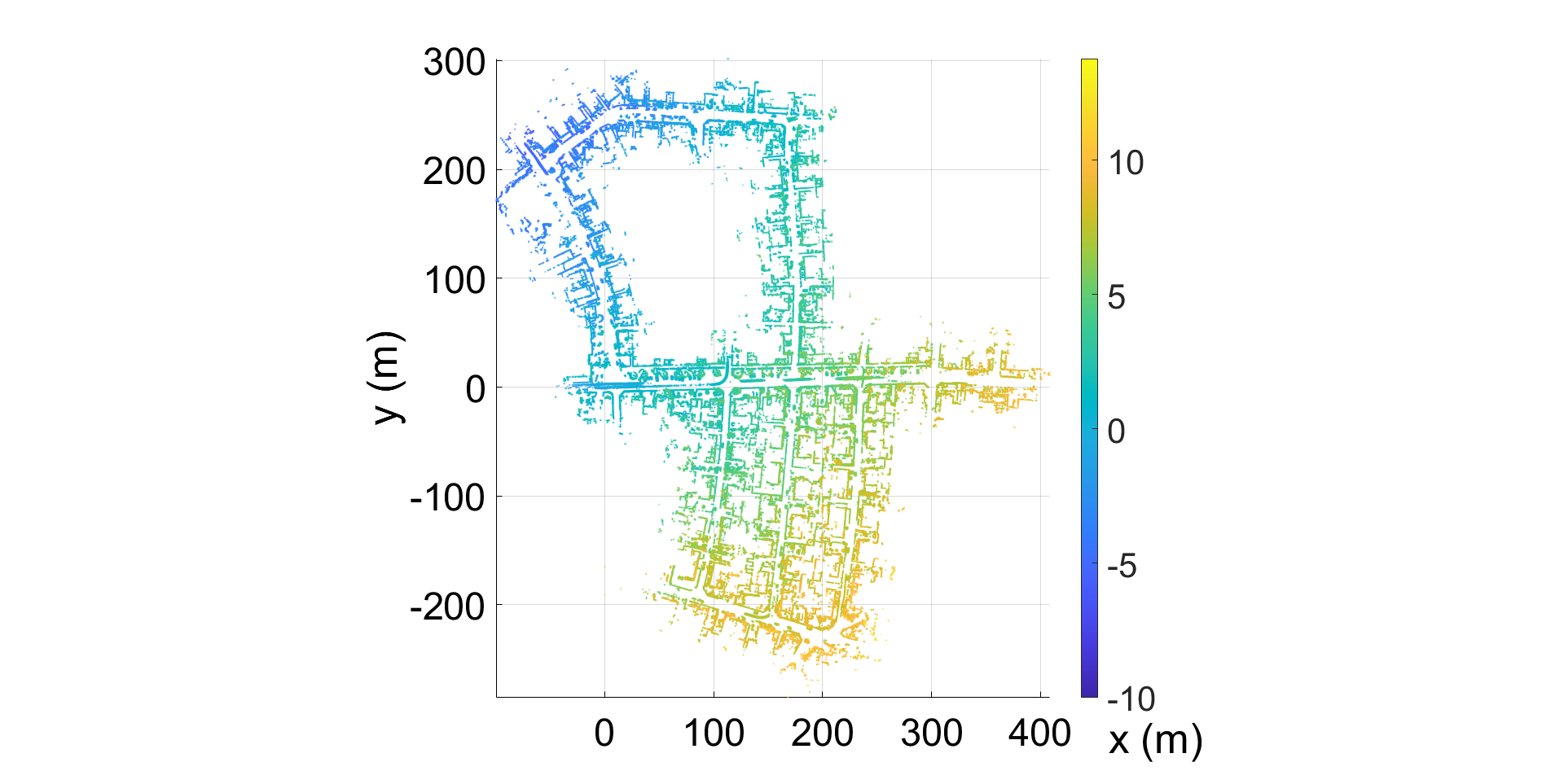}
				\label{pclroadrem4}}
				
	\caption{A sample map from KITTI odometry dataset (sequence 5)}
	\label{pclroadrem}
	\end{figure}

	The scan-to-map matching procedure discussed in section \ref{mrgl} is developed for $2\Dim$ scans. While developing a full $3\Dim$ scan-to-map matching procedure along similar lines is straightforward, the computational complexity renders the approach intractable. To this end, the LIDAR scans are matched to the global map only in $2\Dim$. The success of multi-resolution scan matching method  is highly dependent on distinguishable spatial features in the $2\Dim$ map and the $2\Dim$ LIDAR scans. In the top view of the $3\Dim$ point cloud (by neglecting the z-coordinate), the lowest resolution $2\Dim$ grid will have points almost in every cell. The matching score over road regions will be smaller than the score for grid cells belonging to vertical walls or building. To reduce the number of false positives, it would be advantageous to remove flat regions and hence improve the features in the top view that are dominated by vertical features which are strongly captured by most LIDAR sensors. This is evident in Figure \ref{pclroadrem}. To this end the road segmentation block (RS) removes the roads which contribute to the major flat regions. Once removed, the top view will clearly show the turns and building borders as strong features for matching. Road segmentation methods are well studied and essential for road and lane detection. The approach of conditional random fields  \cite{xiao2018hybrid,rummelhard2017ground} provide effective means for road segmentation. Recently, deep learning models \cite{oliveira2016efficient,caltagirone2017fast} provide alternative ways to segment road regions and subsequently remove them from the map and the scan. In this work, we simply choose to use surface normals to remove points belonging to flat regions. In particular, points with z-coordinate of its normal greater than 0.75 are removed. This approach is computationally efficient and sufficient for enhancing walls and building as shown in Figure \ref{pclroadrem}. The color map in this figure corresponds to the height (z-coordinate) that varies about 20 meters. The dynamical model in equations \eqref{dyneq1}-\eqref{dyneqN} is a $3\Dim$ model which also accounts for the z-coordinate of the vehicle.
	
	\subsection{Scan-to-map matching }\label{mainaprchmatch}
	This block implements a  two stage scan-to-map matching (S2M) process that improves the overall performance of the particle filter. As the scan-to-map matching procedure in section \ref{mrgl} is computationally demanding, it is only applied to the sample/particle with the highest weight.  The assumption is that the particle with the highest weight has a better chance of being close to the the true vehicle state. Let the index of particle with the highest weight be denoted as $j_{max}$. The BEV LIDAR scan (transformed by the  $j^{th}_{max}$ particle pose), with road removed, is first matched to the global BEV map in $2\Dim$ using  the multi-resolution scan matching algorithm of section \ref{mrgl}. To reduce computational time, the match is only performed in a small rectangular region, with side lengths $[L_x,L_y]$, about the the position of the $j_{max}^{th}$ particle $[x^{(j_{max})},y^{(j_{max})}]^T$. The corresponding lower bound and upper bound for the search region of $\mathbf{t}_{(x,y)}$ are taken as $[x^{(j_{max})}-L_{x},y^{(j_{max})}-L_{y}]$ and $[x^{(j_{max})}+L_{x},y^{(j_{max})}+L_{y}]$, respectively. The resultant match, denoted as $[\hat{x}^{(j_{max})},\hat{y}^{(j_{max})},\hat{\theta}^{(j_{max})}]^T$, is accepted only when the match score is greater than the score of the original particle's pose $[x^{(j_{max})},y^{(j_{max})},\theta^{(j_{max})}]^T$.  As the scan is only matched in the $2\Dim$ BEV map, a subsequent match is performed in $3\Dim$  using GICP, which aligns the $2\Dim$ matched scan  with the full $3\Dim$ map. The underlying assumption is that, the aligned pose $[\hat{x}^{(j_{max})},\hat{y}^{(j_{max})},\hat{\theta}^{(j_{max})}]^T$ in $2\Dim$ is close enough to the true $3\Dim$ vehicle pose, so that the GICP algorithm can locally correct the full pose in $3\Dim$. The GICP corrected pose in $3\Dim$  is denoted $\tilde{\xi}^{(j_{max})}_{k}$=$[\tilde{x}^{(j_{max})},\tilde{y}^{(j_{max})},\tilde{z}^{(j_{max})},\tilde{\theta}^{(j_{max})},\tilde{\phi}^{(j_{max})},\tilde{\psi}^{(j_{max})}]^T$ and can be used to directly update the corresponding pose values in $x^{(j_{max})}_k$ to result in the corrected particle state $\tilde{x}_k^{(j_{max})}$. 
	
\begin{figure}
		\centering
	
		\subfigure[L6 resolution]{\includegraphics[trim=14.5cm 0cm 14cm 0cm,clip=true,width=2.3in]{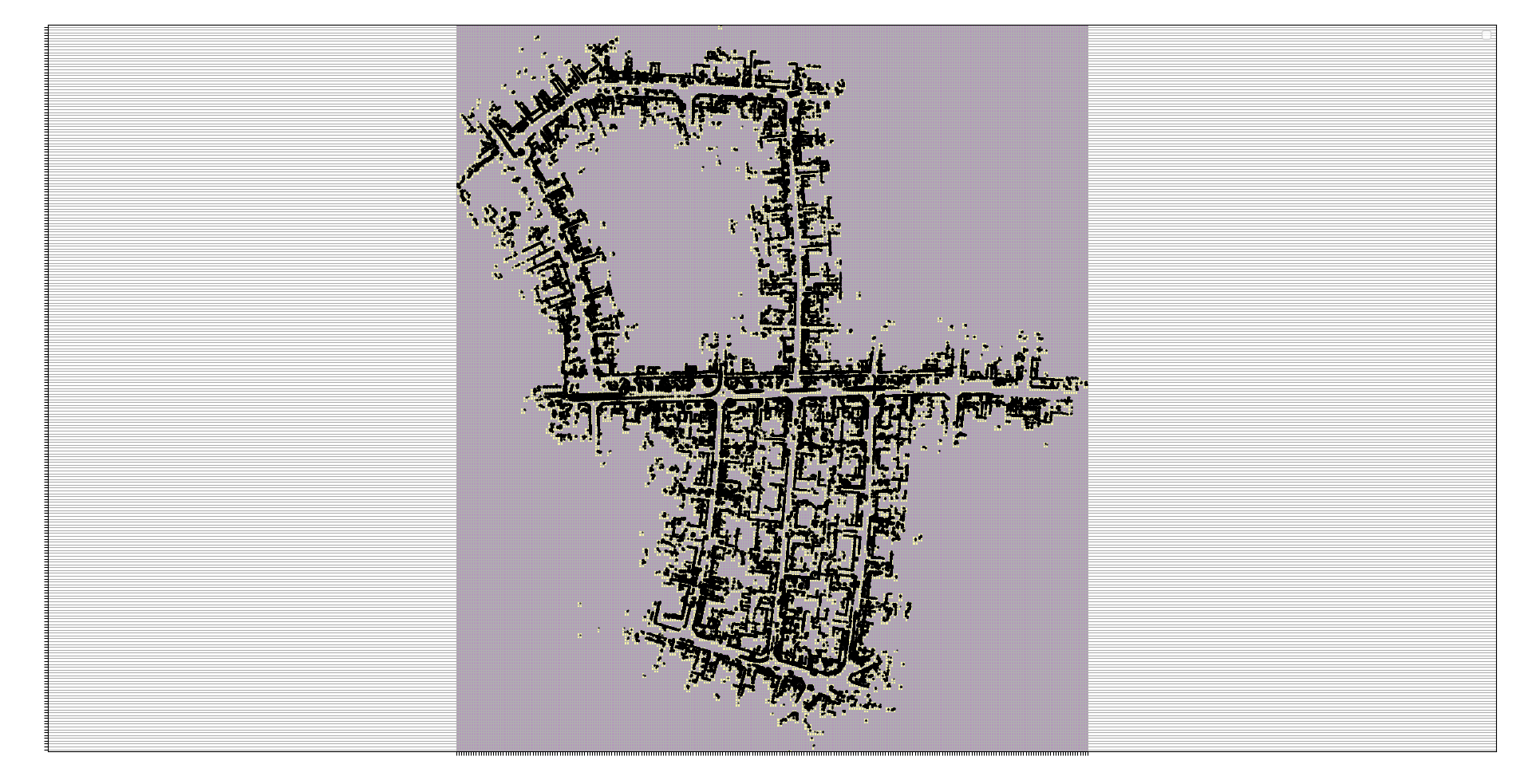}
		\label{multires0}}
	
		\subfigure[L7  resolution]{\includegraphics[trim=14.5cm 0cm 14cm 0cm,clip=true,width=2.3in]{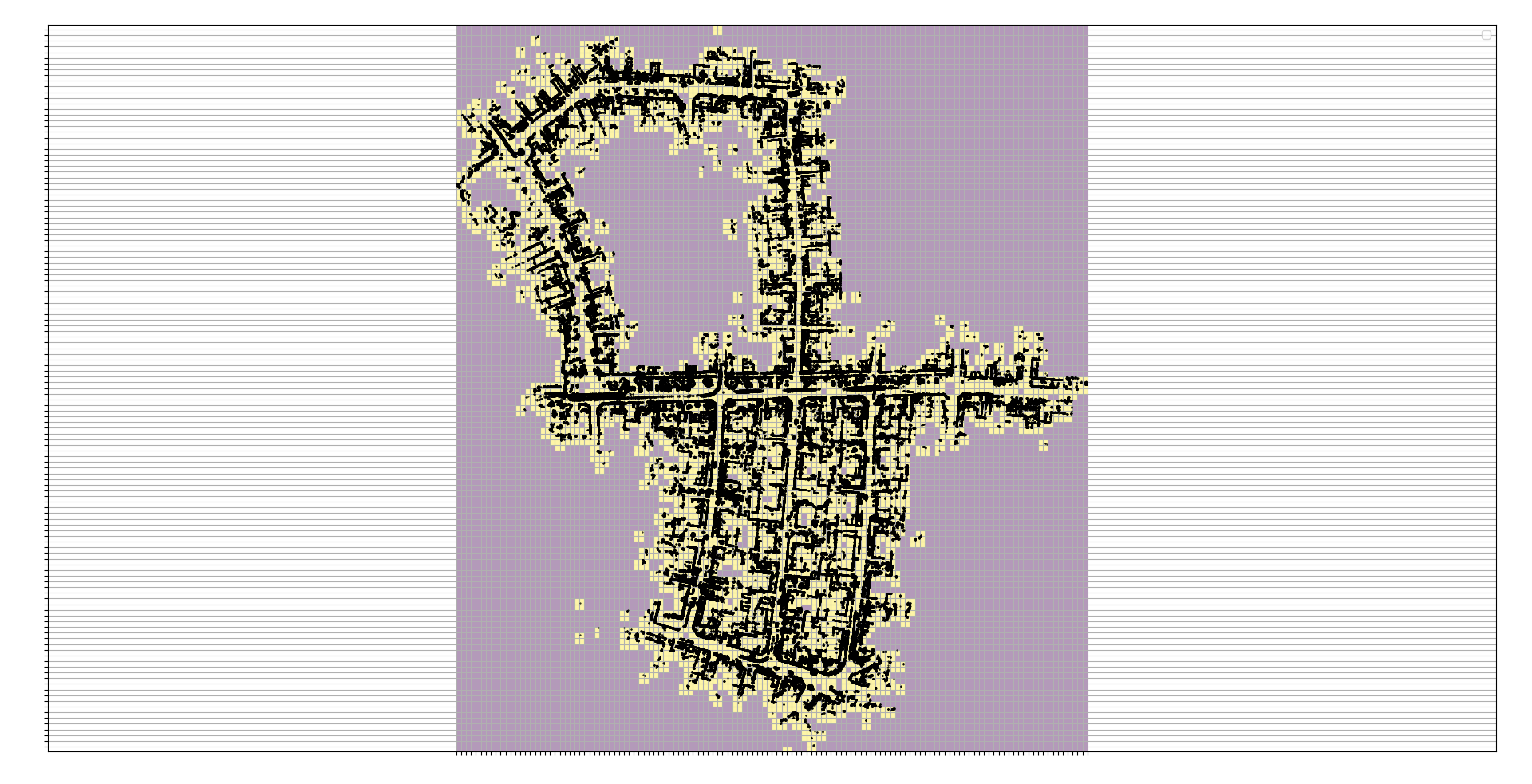}
		\label{multires1}}

	\subfigure[L8 resolution]{\includegraphics[trim=14.5cm 0cm 14cm 0cm,clip=true,width=2.3in]{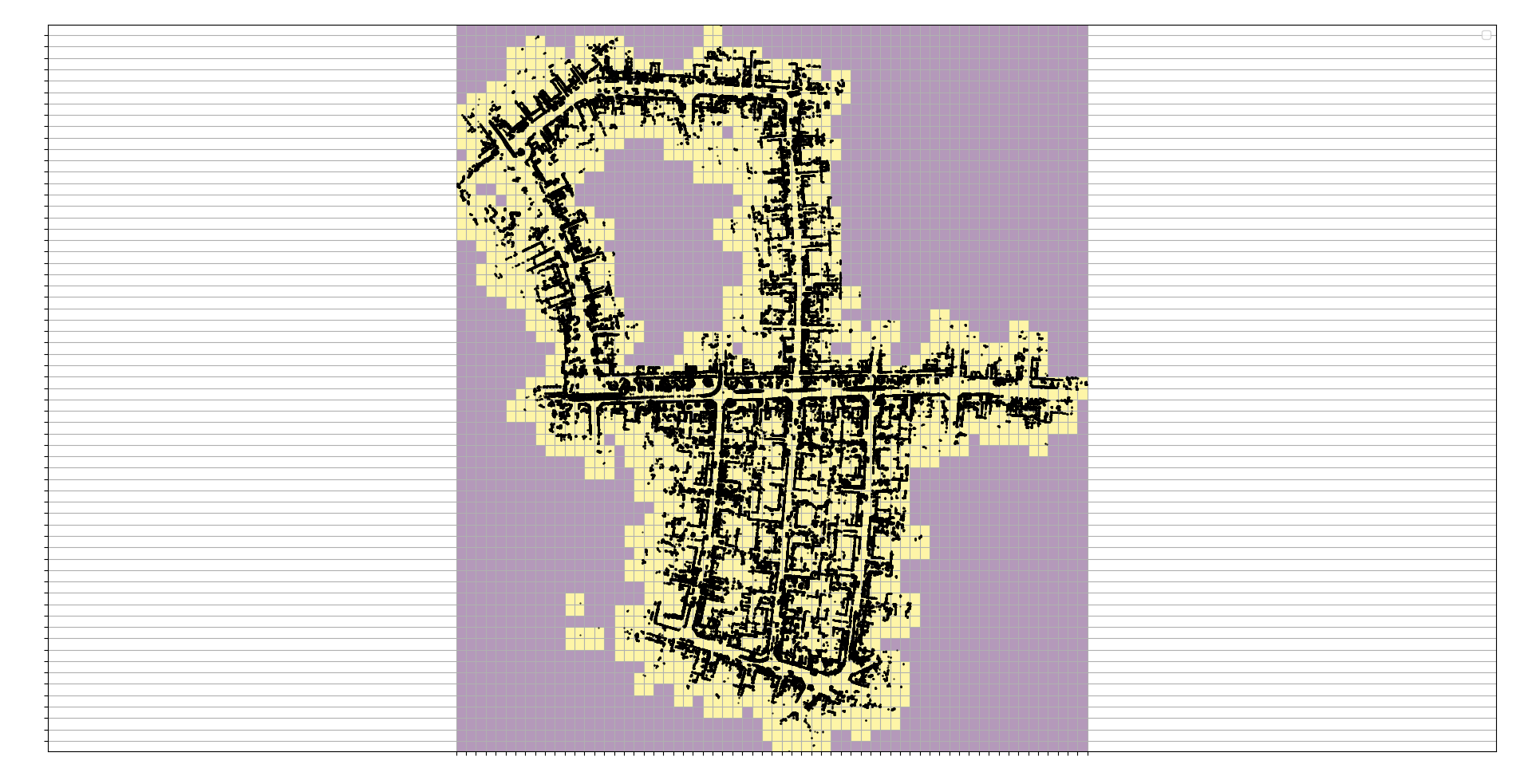}
	 \label{multires2}}

		\subfigure[L9 resolution]{\includegraphics[trim=14.5cm 0cm 14cm 0cm,clip=true,width=2.3in]{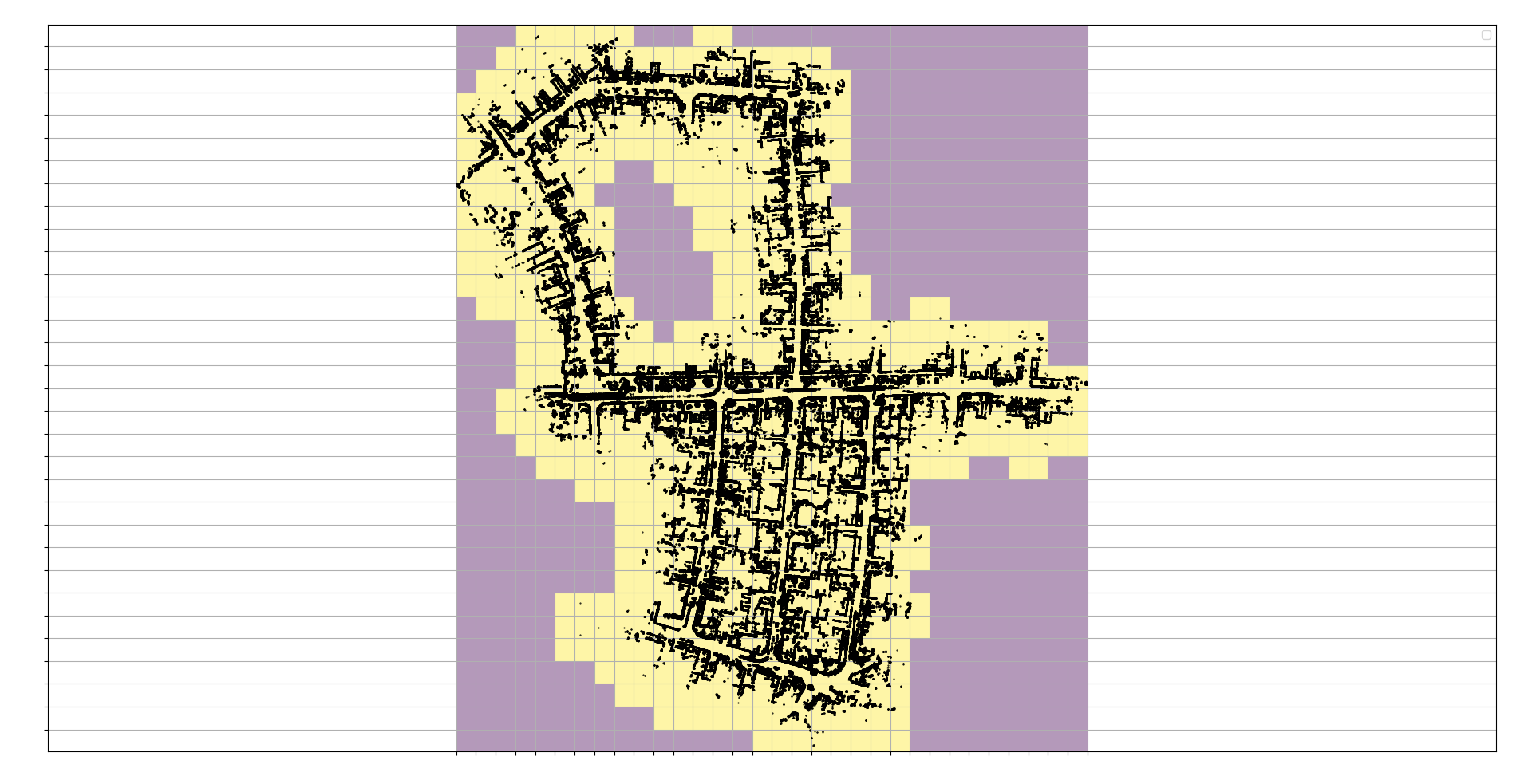}
		\label{multires3}}
	
	\caption{L9 resolution} \label{fullmapresbev}
	\end{figure}

\begin{figure}
			\centering
			\includegraphics[trim=0cm 5cm 5cm 0cm,clip=true,width=5in]{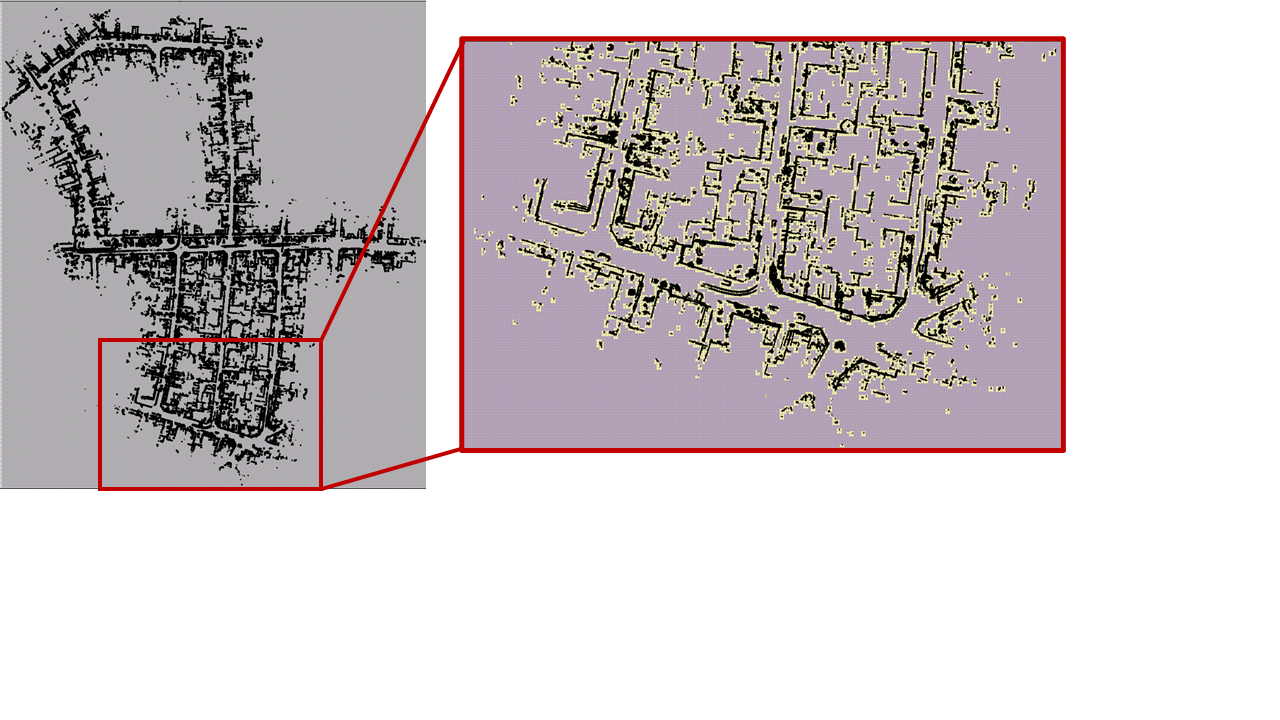}
			\caption{Level $L_{4}$ resolution with zoomed in view at a particular region} \label{fullmapresbev2}
		%
	\end{figure}

	The multi-resolution levels for the map are pre-computed offline and directly used during the S2M block computations. For example, Figure \ref{pclroadrem4} shows the BEV map, of sequence '05' of the KITTI odometry dataset,  with horizontal planes removed. The KITTI dataset also contains true ground pose data that is used to assemble all the individual LIDAR scans and form the map $\mathcal{M}$. The corresponding $2\Dim$ map $\mathcal{M}_{bev}$ is constructed by neglecting the z-coordinate. Few instances of the corresponding  multi-resolution grids are shown in Figure \ref{fullmapresbev} for levels $L6$-$L9$. In this figure, the black points indicate the LIDAR scan points of the map, the lighter yellow regions represent the grid cells that have at least one map point. The level with highest resolution $L0$ (not shown in figure) was constructed with the desired matching resolution of $d_{match}=2m$. The subsequent levels were constructed by the max pool operation as described in section \ref{mrgl}. Figure \ref{fullmapresbev2} also shows the zoomed in image of the $L4$ resolution grid. It can be seen that by removing the horizontal surfaces (roads and roofs of building), the LIDAR features for matching become prominent.  


\begin{figure}
			\centering
	
			\subfigure[RMSE histogram]{\includegraphics[width=3in]{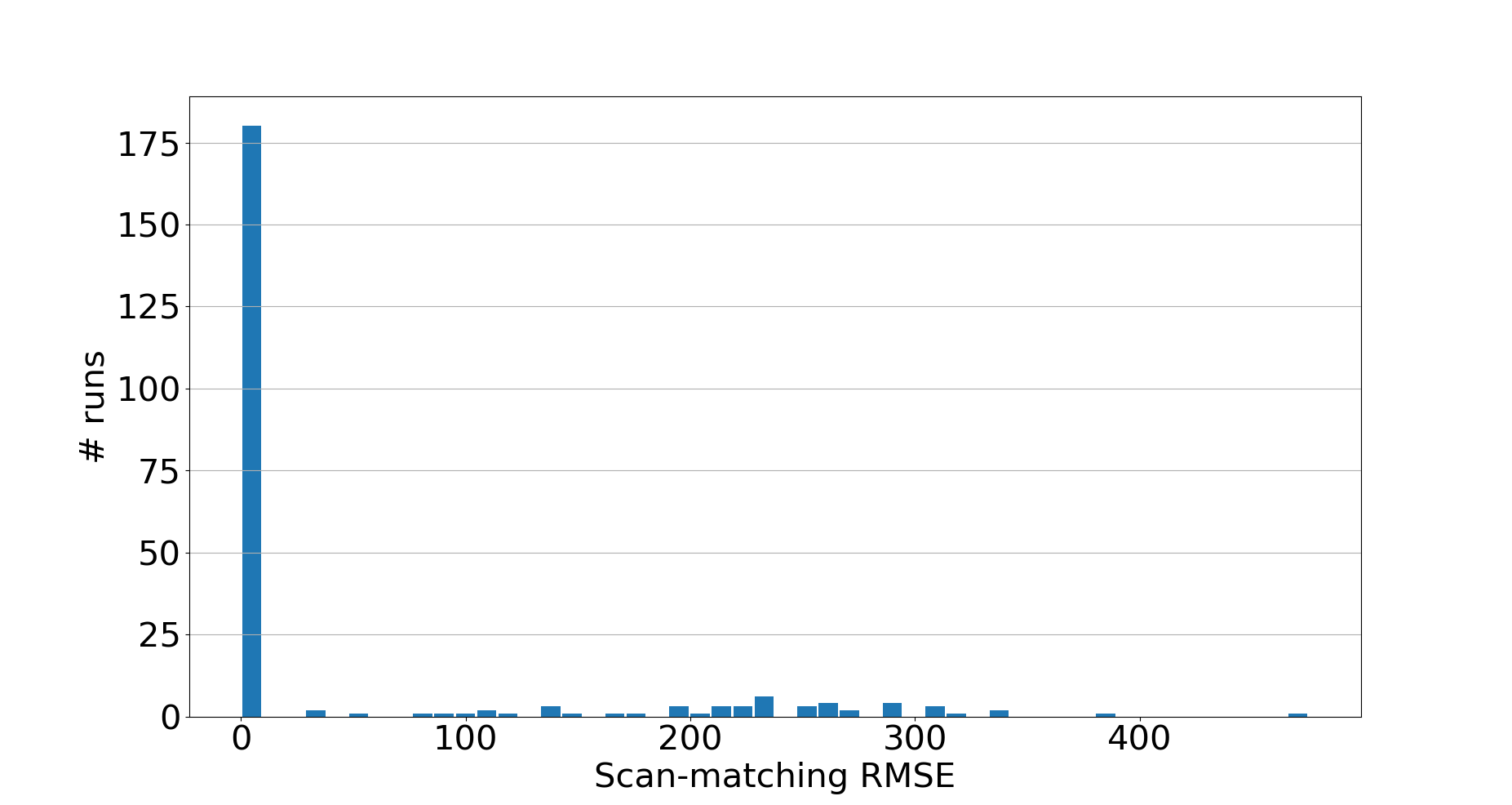}
			 \label{scanmatch5001}}

			\subfigure[RMSE histogram for $rmse\le 2.5$]{\includegraphics[width=3in]{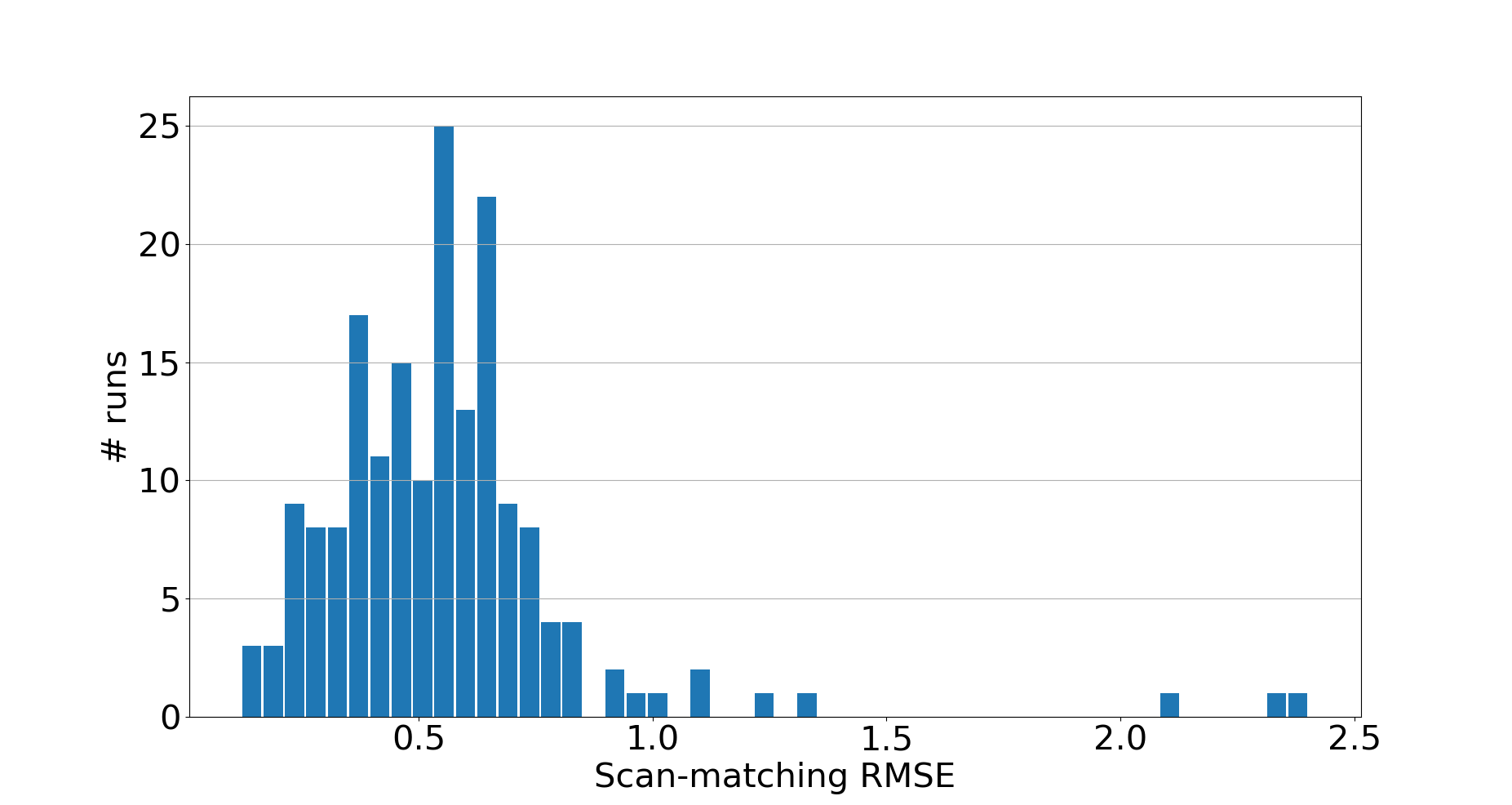}
			 \label{scanmatch500cropped}}

		\caption{RMSE histogram for scan-to-map matching  (x-axis units in meters)}
		\label{scanmatch500}
		
	\end{figure}

	Figure \ref{scanmatch5001} shows the Root-Mean-Squared-Error (RMSE) results of matching individual LIDAR scans from the KITTI odometry dataset sequence-05 with the corresponding map $\mathcal{M}_{bev}$. The histogram is constructed by randomly sampling 350 individual scans from the complete dataset of LIDAR scans of the sequence-05. This scan is matched to the map $\mathcal{M}_{bev}$ using the multi-resolution scan matching approach discussed in section \ref{mrgl}. The RMSE is calculated using the euclidean distance between the $2\Dim$ scan points transformed by the true pose and the corresponding $2\Dim$ points transformed by the computed pose.  The angular resolution was taken as $d_{\theta}=2.5\deg$ in the angular range of $[-\pi,\pi]$, the matching resolution was taken as $d_{m}=[1m,1m]$, and the search space for the translation was taken as the complete range of the map $\mathcal{M}_{bev}$ as in Figure \ref{pclroadrem4}, which is about 400m in both x and y dimensions. Matching a scan against this full map is computationally demanding and time consuming. Average wall-clock time for computing the match is 1.6 minutes, which is not acceptable for real-time performance. Yet the results show strong optimism in matching performance, with majority of the RMSEs close to zero.  Figure \ref{scanmatch500cropped} shows the same histogram results but zoomed in the interval $[0m,2.5m]$, the RMSE is within the chosen matching resolution of 1m. It can also be seen that the results in Figure \ref{scanmatch5001} show large RMSE for some scans, indicating failed matches. This is often the case when a LIDAR scan is matched against the entire map with multiple sections that look very similar, especially at low resolution levels.  But, it can also be observed that the scan-to-map matching succeeds for majority of the cases ($>87\%$).  We use OpenMP to parallelize the computation of the matches and use the priority queue from the C++ Standard Template Library (STL)  to implement  max heap for the matching algorithm. Note that, the implementation may not be optimized for execution time. GPU implementation can make the scan matching much faster.   
	The computational time and the failed matches can be significantly reduced by searching in a local neighborhood $[L_x,L_y]$ rather than the full map. This reduces the search space for the translation variable $\mathbf{t}_{(x,y)}$. The computational complexity can also be reduced by using a down-sampled scan.

	\subsection{Particle state correction}\label{mainaprchjmax} 
	As seen in the previous section the scan matching process is often computational expensive than the particle filter, which only entails simple likelihood computations for each particle. While the ideal case would be to apply scan-to-map matching (section \ref{mrgl}) for every particle, it would be computationally prohibitive and unnecessary. The particle filter thread (in Figure \ref{flchart}) often runs at higher frame rate (limited by LIDAR scan frame rate) compared to the slower frame rate of scan-to-map matching thread. This might make the scan-to-map matching thread useless for any real-time implementation. Despite the slow frame rate of the scan matching process, its high success rate in exactly matching the scan to the map is highly advantageous, and should be leveraged. To this end, we propose a two stage scan-to-map matching process that accounts for the slower frame rate. In the first stage, only the pose for the $j_{max}$ particle at some time step $t_k$ is used to begin the scan matching process. When the matching process completes at some later time step $t_{k'}$, the second stage deals with the process of propagating the matched pose from time step $t_k$ to $t_{k'}$. To improve the success of matching the scan to the map and to reduce the computational time, only the pose of the $j_{max}$ particle is matched to the map. The scan-to-map matching process that is initiated at time step $t_k$ with pose ${\xi}^{(j_{max})}_{k}$ of the $j_{max}$ particle,  will result in the corrected pose $\tilde{\xi}_k^{(j_{max})}$ at some later time step $t_{k'}$ where $t_{k'}=t_{k}+n_{s2m}\Delta t$, with $n_{s2m}\Delta t$ being the number of time steps required by the the S2M procedure. One might be inclined to simply correct the pose $\xi^{(j_{max})}_{k'}$ of the same $j_{max}$ particle at time step $t_{k'}$ with the S2M pose $\tilde{\xi}^{(j_{max})}$ computed at time $k$. While this might be pragmatic when the vehicle is moving slowly,  most often this will lead to a low likelihood weight for the particle at time $t_{k'}$ as the vehicle could have moved a significant distance or could have made a significant turn, thus wasting all the computations of the S2M block. The appropriate solution would be to propagate the computed S2M pose $\tilde{\xi}^{(j_{max})}_{k}$ at time $t_k$ to the current time $t_{k'}$. This is done using relative poses $H_{(t_{k'},t_k)}$ computed from the RPE block from time $t_k$ to $t_{k'}$ as 
	\begin{align}
	H(\tilde{\xi}^{(j_{max})}_{k'})=H_{(t_{k'},t_k)}H(\tilde{\xi}^{(j_{max})}_{k})
	\end{align}
	where $H(\tilde{\xi}^{(j_{max})}_{k})$ is the $4\times 4$ homogeneous transformation matrix corresponding to the pose $\tilde{\xi}^{(j_{max})}_{k}$. $\tilde{\xi}^{(j_{max})}_{k'}$ can now be used to update the pose coordinates of  $x^{(j_{max})}_{k'}$ at time $t_{k'}$. Further, the velocity $\|\bar{v}_{k'}\|$ and angular rate $\|\dot{\bar{\zeta}}_{k'}\|$ estimates from the RPE block are also used to directly replace the corresponding values in $x^{(j_{max})}_{k'}$.  During the time $[t_k,t_{k'}]$ no re-sampling is performed to keep the $j_{max}$ particle intact. The key assumption of this block is that the relative pose estimates during the duration $n_{s2m}\Delta t$ can be computed reliably and accurately using the GICP algorithm of the RPE block. This allows one to use  relative transformation $H_{(t_{k'},t_k)}$ to transform the accurate global pose correction at time $t_{k}$ to $t_{k'}$.

	\section{Simulation}\label{sims}
	In this section, the proposed approach is validated on the publicly available KITTI odometry dataset. These datasets have ground truth poses that allow for checking convergence and accuracy of the localization and tracking processes.  As with most algorithms, the proposed approach is also dependent on  tuning parameters such as $N_p$, $Q_k$, $d_{max}$, $\sigma$, resolution for voxel-based downsampling. As it is well established that a very large number of particles will improve convergence of the particle filter  \cite{blanco2019benchmarking}, we only use $N_p=1000$ particles in our preliminary simulations to test the performace of localization using fewer particles. The simulations help validate the hypothesis that scan-to-map matching can considerably increase convergence of the particle filter with just few particles.  We fix the resolution for downsampling at 0.5m for the LIDAR scans, $d_{max}$ is taken as $5m$ and $\sigma=0.5m$. The process noise for the state $\mathbf{x}$ in the dynamical model, described in section \ref{pfsecdynmodel}, is taken as $Q_k=diag[0.5^2,0.5^2,0.1^2$, $(5^o)^2,(2^o)^2,(0.1^o)^2$, $0.1^2$, $(0.05^o)^2,(0.05^o)^2,(0.02^o)^2,0.001^2]$, where $diag[.]$ represents a diagonal matrix with given entries along the diagonal. The resolution of voxel-downsampling for the LIDAR scan can be chosen to achieve the desired frame rate of the particle filter thread. For the particle filter, the initial particles have positions uniformly sampled in the entire $3\Dim$ map region, the yaw-pitch-roll angles are sampled from $[-180\deg,180\deg]$,$[-30\deg,30\deg]$ and $[-10\deg,10\deg]$ respectively .The angular rates sampled from a Gaussian probability density function (PDF) with zero mean and variance $(0.1\:rad/s)^2$. The velocity is assumed to be uniformly sampled from $[0,v_{max}]$, where $v_{max}$ is the known maximum velocity for the vehicle ( used in sequence 05 of KITTI dataset \cite{Geiger2012CVPR}) or based on the location within the map (speed limits) or the average speed of the vehicle.   
	 
The simulation implements the parallel architecture of the particle filter and the scan-to-map matching procedure as illustrated in 	Figure \ref{flchart}.  As the particle filter is  initiated with random particles over the entire $3\Dim$ map, only few particle or none of the particles will be close to the true position of the vehicle. As the vehicle moves around in the map, the sequential LIDAR scans are processed by the particle filter thread block and the RPE thread block to estimate the vehicle's state.  Particle close enough to the true vehicle state will eventually be weighted higher by the particle filter. Hence, to achieve successful implementation of the particle filter, large number of particles are required, especially in $3\Dim$, to improve the chances of having particles close enough to the the true vehicle state. The S2M block tries to place particles closer to the vehicle state and thus the particle filter is able to use fewer particles (just 1000 particles in the simulations). In essence, the computational complexity of using large number of particles in a convetional particle filter is reduced by using the S2M block with fewer particles.  

\begin{figure}
\centering
\includegraphics[trim=13cm 2cm 11cm 1cm,clip=true,width=4in]{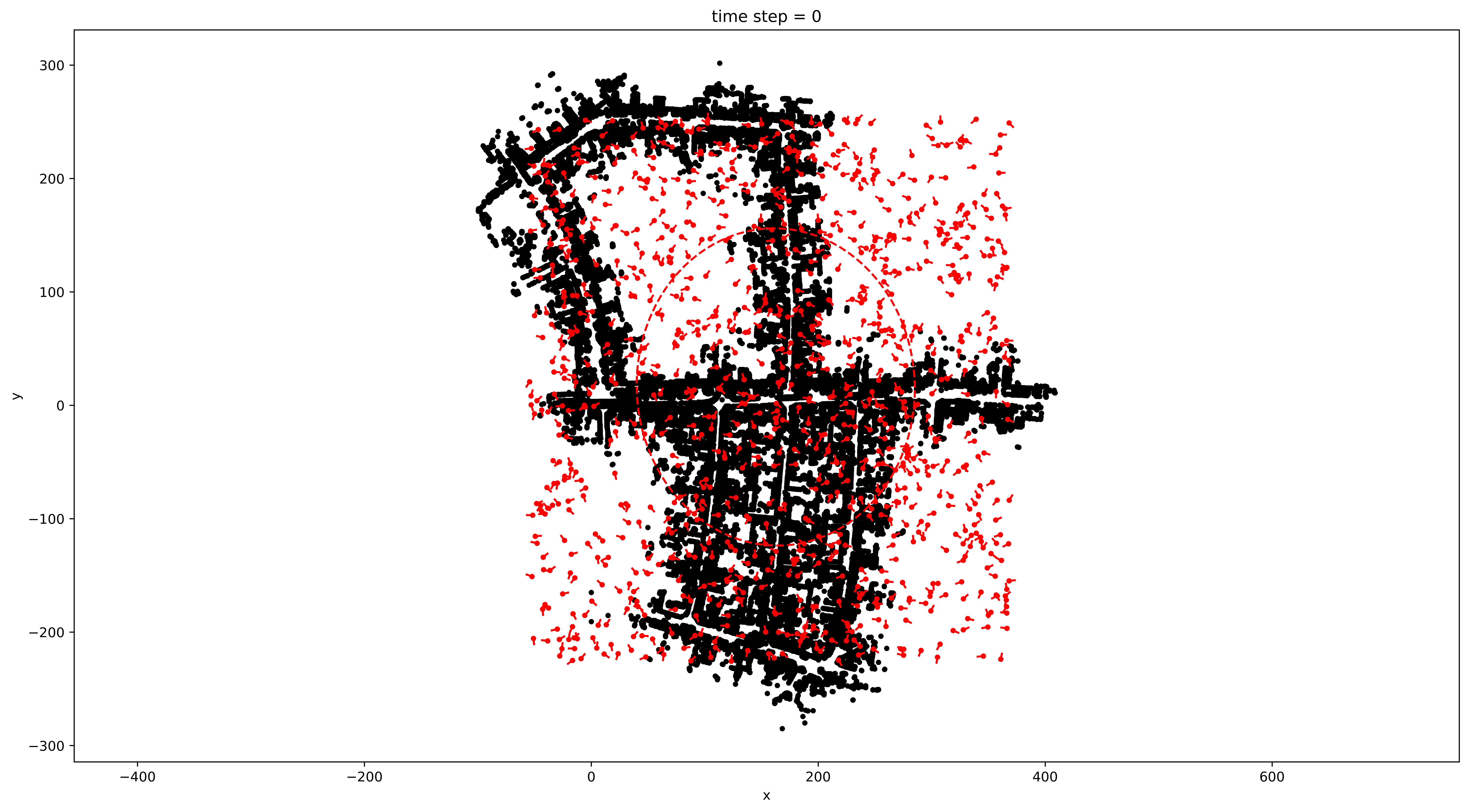}
			\caption{Snapshot at $k=0$ with particles shows as red dots} \label{snapshoteg1_0}
\end{figure}
Figure \ref{snapshoteg1_0} shows the initial time step where the particle filter is initialized over the whole map. The figure shows a $2\Dim$ top view or BEV of the $3\Dim$ map. While the positions of the particles are initialized in $3\Dim$, only the $2\Dim$ projection are shown in Figure \ref{snapshoteg1_0} as red circular markers (or dots) with a small line segment indicating the heading of the particle. As observed all the particles have random heading and positions, and thus most particles have very low likelihood. Only particles close enough to the true position of the particle will gain higher weights through higher relative likelihood values. But with only 1000 particles, the probability of a particle to fall close to the true position and also have close enough heading (pitch and roll) to the true vehicle is very small. The only solution is to have massive number of particles, especially in $3\Dim$, to be close to the vehicle's pose, which in turn makes the particle filter inefficient and even computationally intractable. The dotted red circle in the figure corresponds to the $2\sigma$ covariance ellipsoid of the $[x,y]$ position estimated by the particle filter.

\begin{figure}
\centering

			\subfigure[Full map]{\includegraphics[trim=10cm 2cm 9cm 1cm,clip=true,width=2.6in]{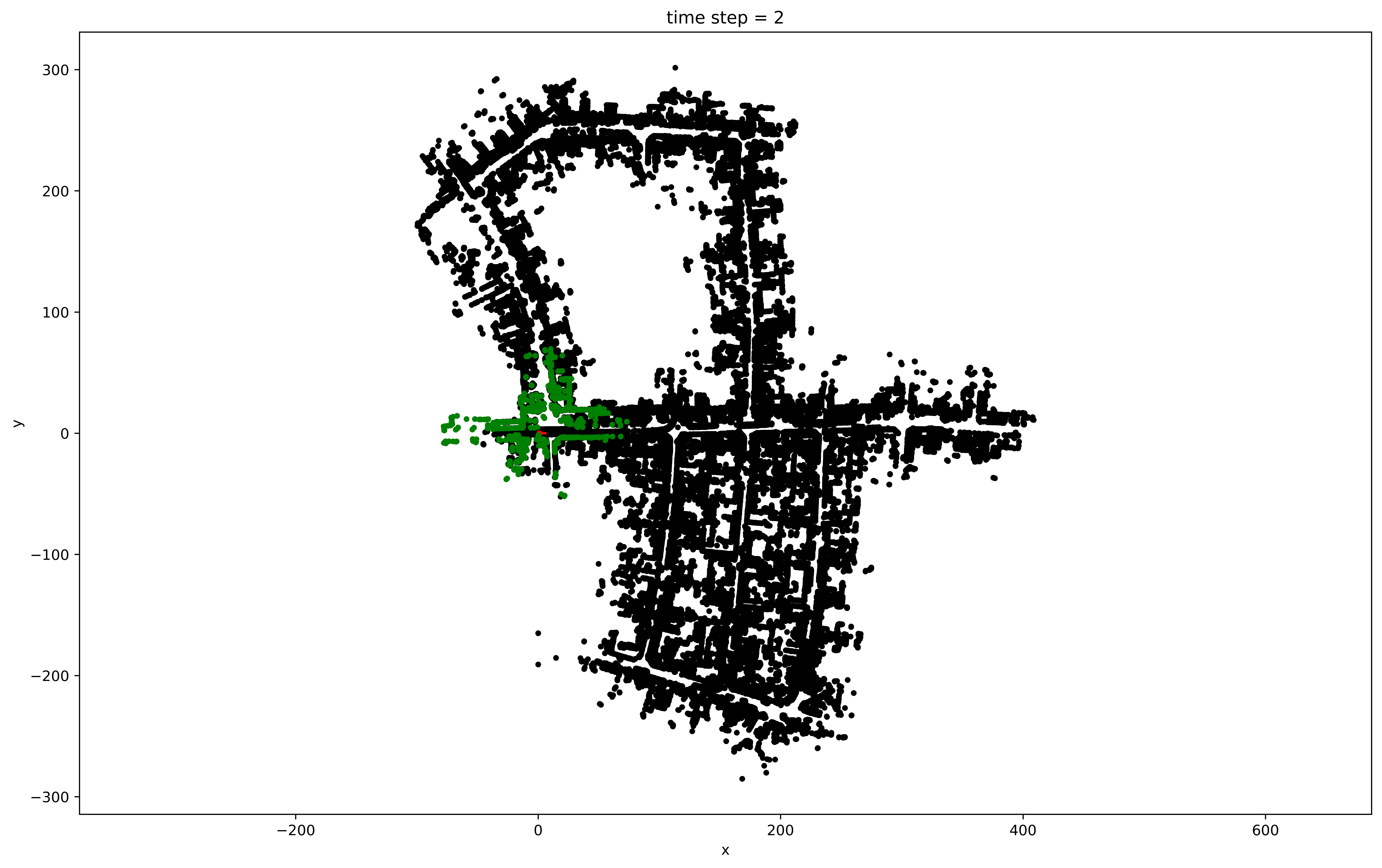}
			 \label{snapshoteg1_2}}

			\subfigure[Zoomed in image]{\includegraphics[trim=10cm 2cm 8cm 1cm,clip=true,width=2.5in]{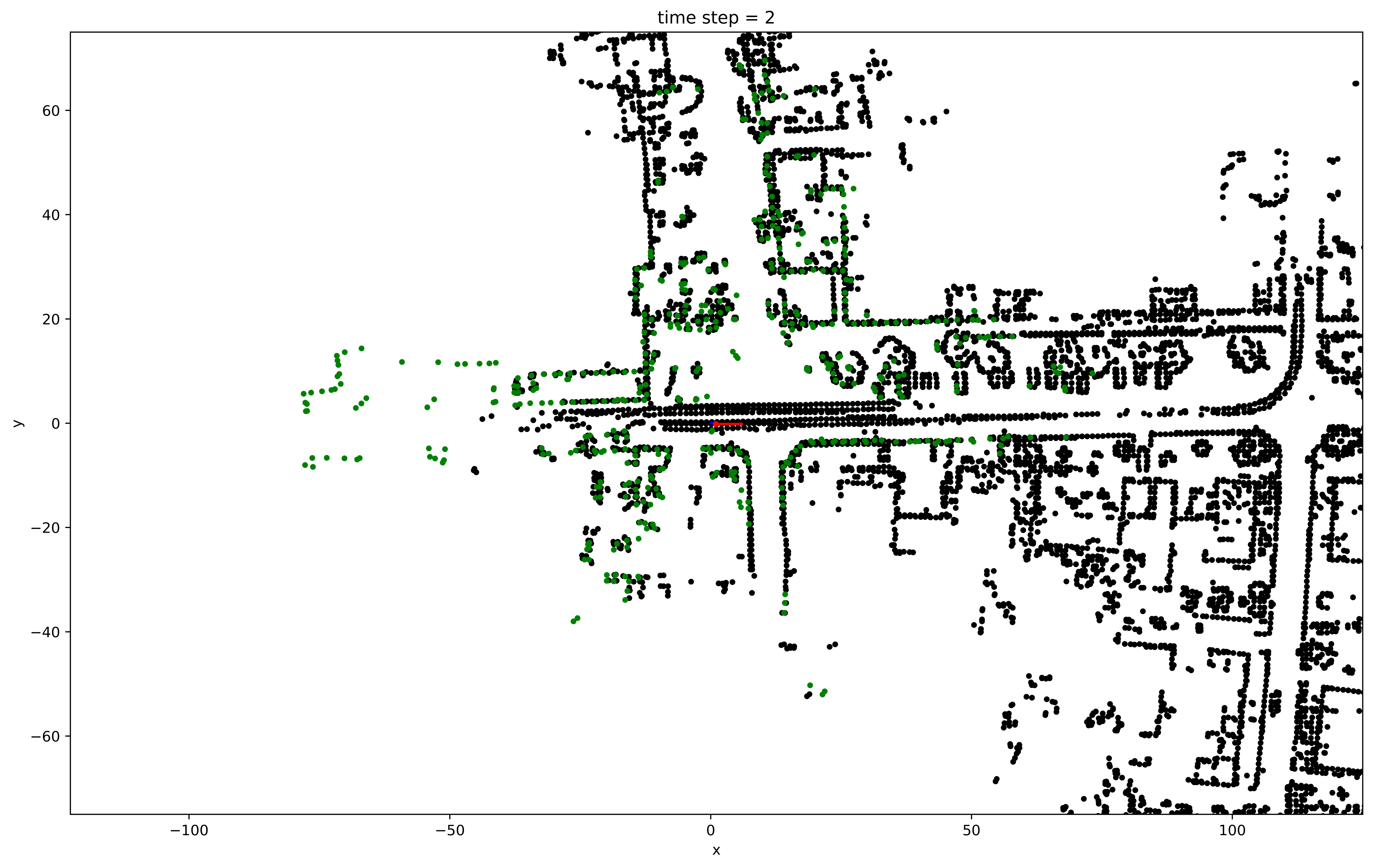}
			 \label{snapshoteg1_3}}

	\caption{Snapshots of a simulation at $k=2$}
		\label{snapshots1a}
\end{figure}

Figure \ref{snapshots1a} shows the snapshot of the simulation and its corresponding zoomed in region (Figure \ref{snapshoteg1_3}) around the initial position at time step $k=2$. At time step $k=0$, the S2M thread begins the scan-to-map matching process for the particle with the highest weight. At $k=2$, the S2M results in a match for the LIDAR scan at $k=0$. The matched pose is then propagated to the current time step $k=2$. This matched and propagated pose is used to transform the LIDAR scan at time step $k=2$ from the body-fixed frame to the initial (inertial) frame and is shown using green points in Figure \ref{snapshots1a}. It can be observed that the green points overlap the black map points, indicating a very good match. This suggests that the LIDAR scan at time step $k=0$ was matched accurately and the relative pose estimates were good enough to propagate the exact match to time step $k=2$. Once the matched pose estimates are used to update the particle with the highest weight, the likelihood of this particle is recomputed to assign new a weight that is typically the highest on account of a very good match. As a result, the resampling process of the particle filter will resample the particles close to this particle. This can be observed in zoomed Figure \ref{snapshoteg1_3}, where the particles represented by red points are overlapped. S2M thread again begins at this time step and completes at some later time step ($k>2$).     

\begin{figure}
\centering
	
			\subfigure[Full map]{\includegraphics[trim=13cm 2cm 11cm 1cm,clip=true,width=2.6in]{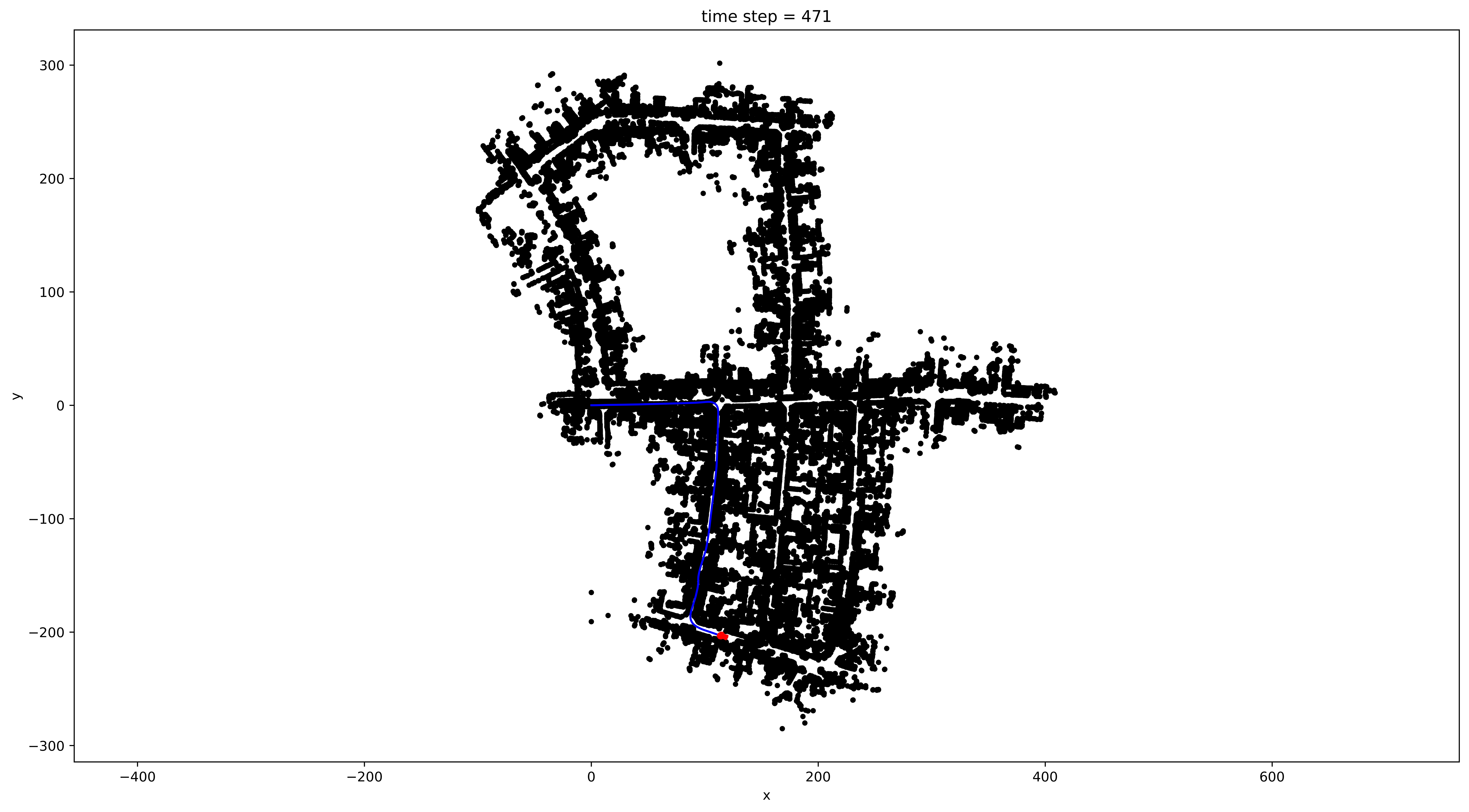}
			 \label{snapshoteg1_4}}

			\subfigure[Zoomed in image ]{\includegraphics[trim=10cm 2cm 8cm 1cm,clip=true,width=2.5in]{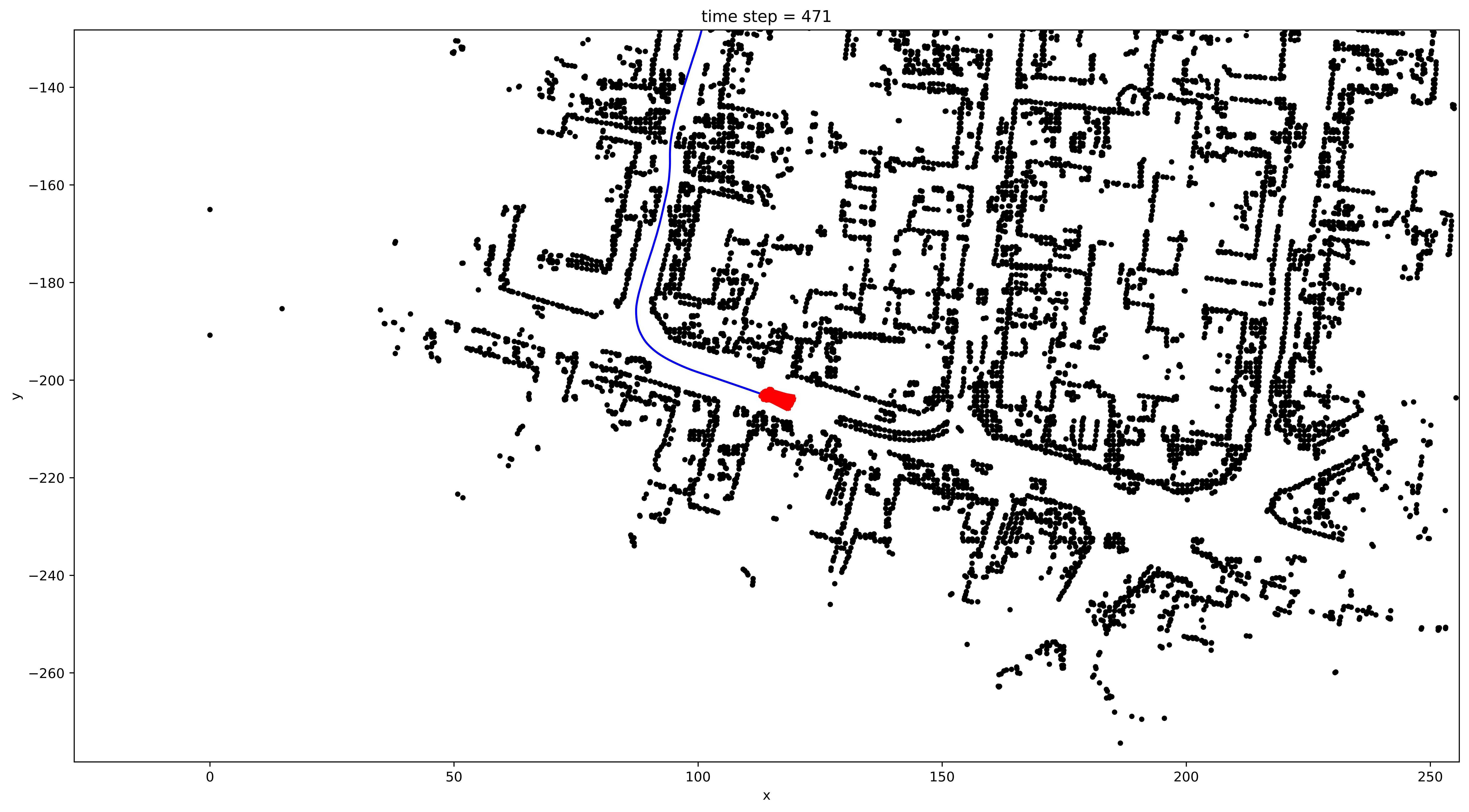}
			 \label{snapshoteg1_5}}

\caption{Snapshots of a simulation at $k=471$}
		\label{snapshots1b}
\end{figure}

Figure \ref{snapshots1b} shows similar snapshots of the simulation at time step $k=471$. The blue line is the ground truth trajectory of the vehicle. At this time step there is no result from the S2M thread yet, in which case the conventional particle filter keeps track of the vehicle's state.  Figure \ref{snapshoteg1_5} shows that all the particles are close together at the true position of the vehicle, thus providing reliable estimates between S2M thread computations. As such, the particle with the highest weight is often close to the vehicle's state thus leading to faster and successful S2M thread results. 
  
\begin{figure}
\centering

			\subfigure[Full map]{\includegraphics[trim=13cm 2cm 11cm 1cm,clip=true,width=2.6in]{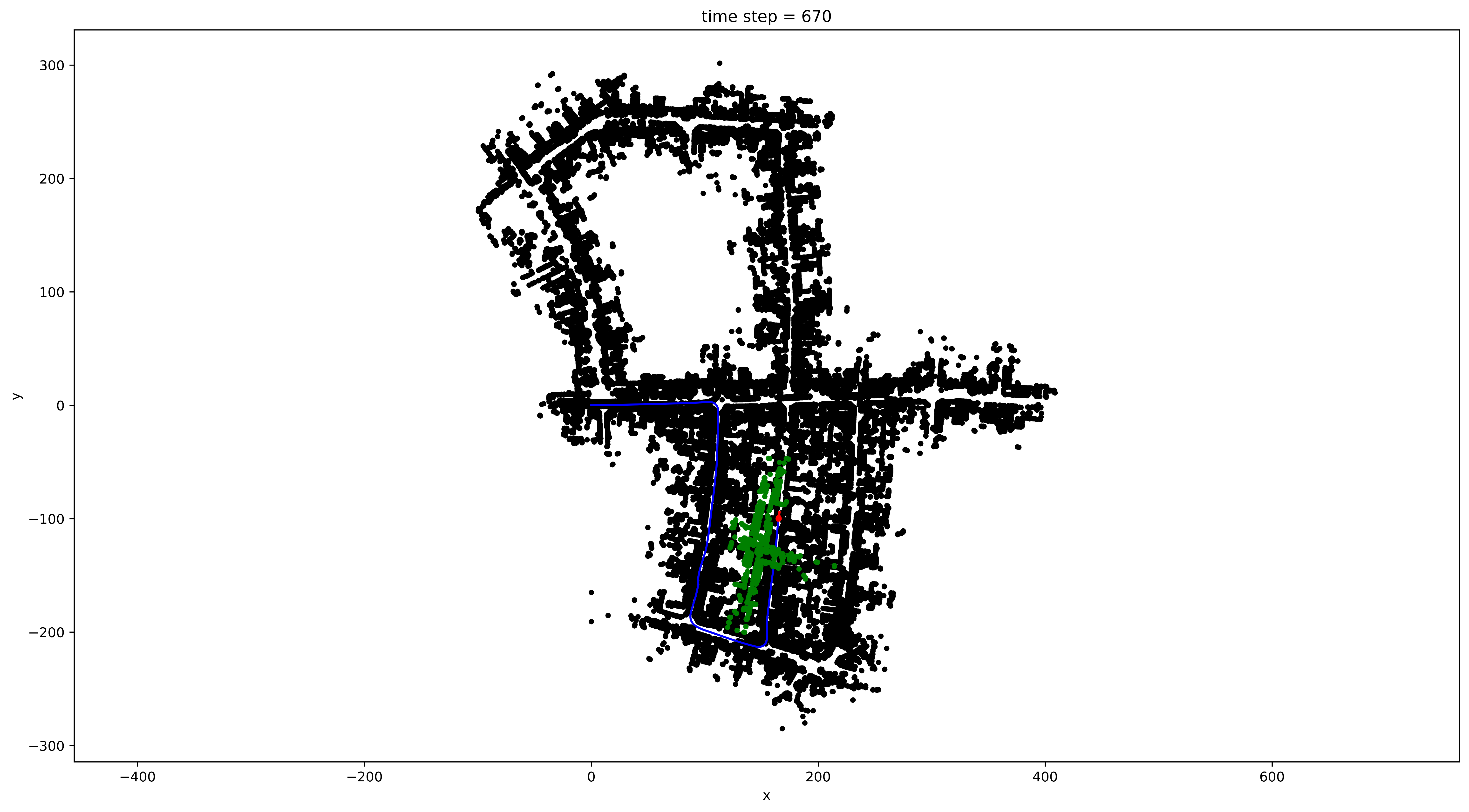}
			\label{snapshoteg1_6}}

			\subfigure[Zoomed in image ]{\includegraphics[trim=10cm 2cm 8cm 1cm,clip=true,width=2.5in]{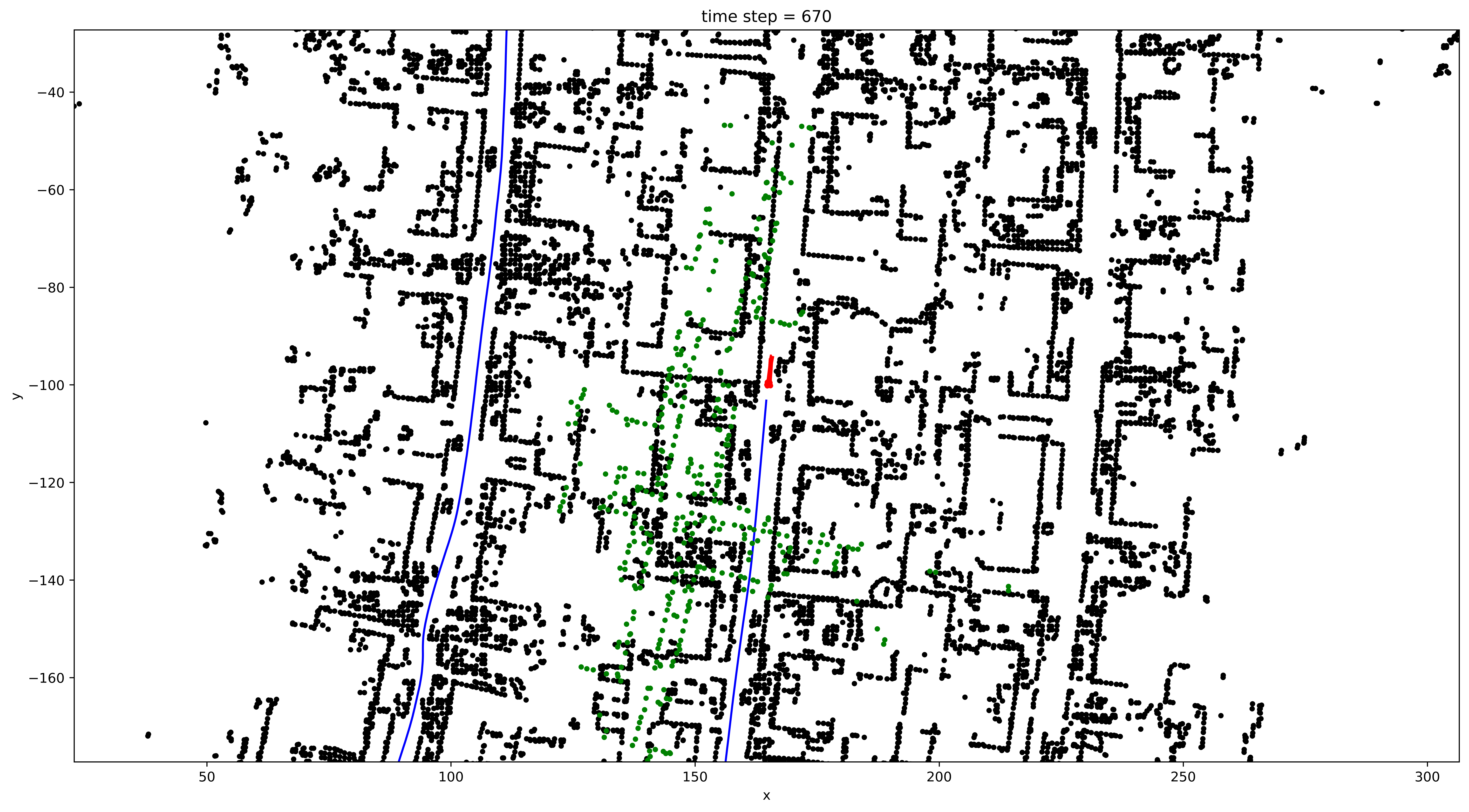}
			 \label{snapshoteg1_7}}

\caption{Snapshots of a simulation  at $k=670$}
		\label{snapshots1c}
\end{figure}
\begin{figure}
\centering

			\subfigure[Full map ]{\includegraphics[trim=13cm 2cm 11cm 1cm,clip=true,width=2.6in]{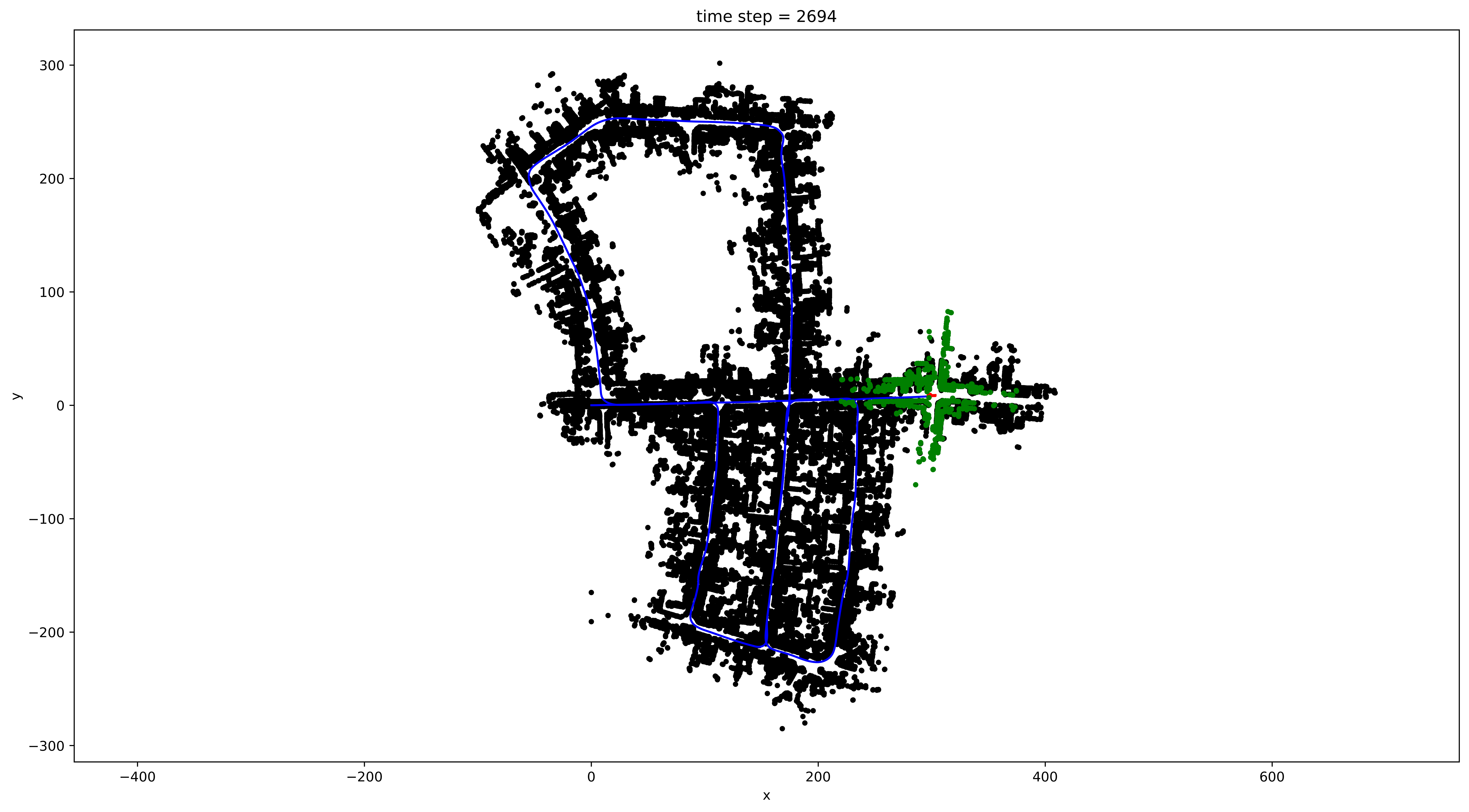}
			\label{snapshoteg1_8}}

			\subfigure[Zoomed in image ]{\includegraphics[trim=10cm 2cm 8cm 1cm,clip=true,width=2.5in]{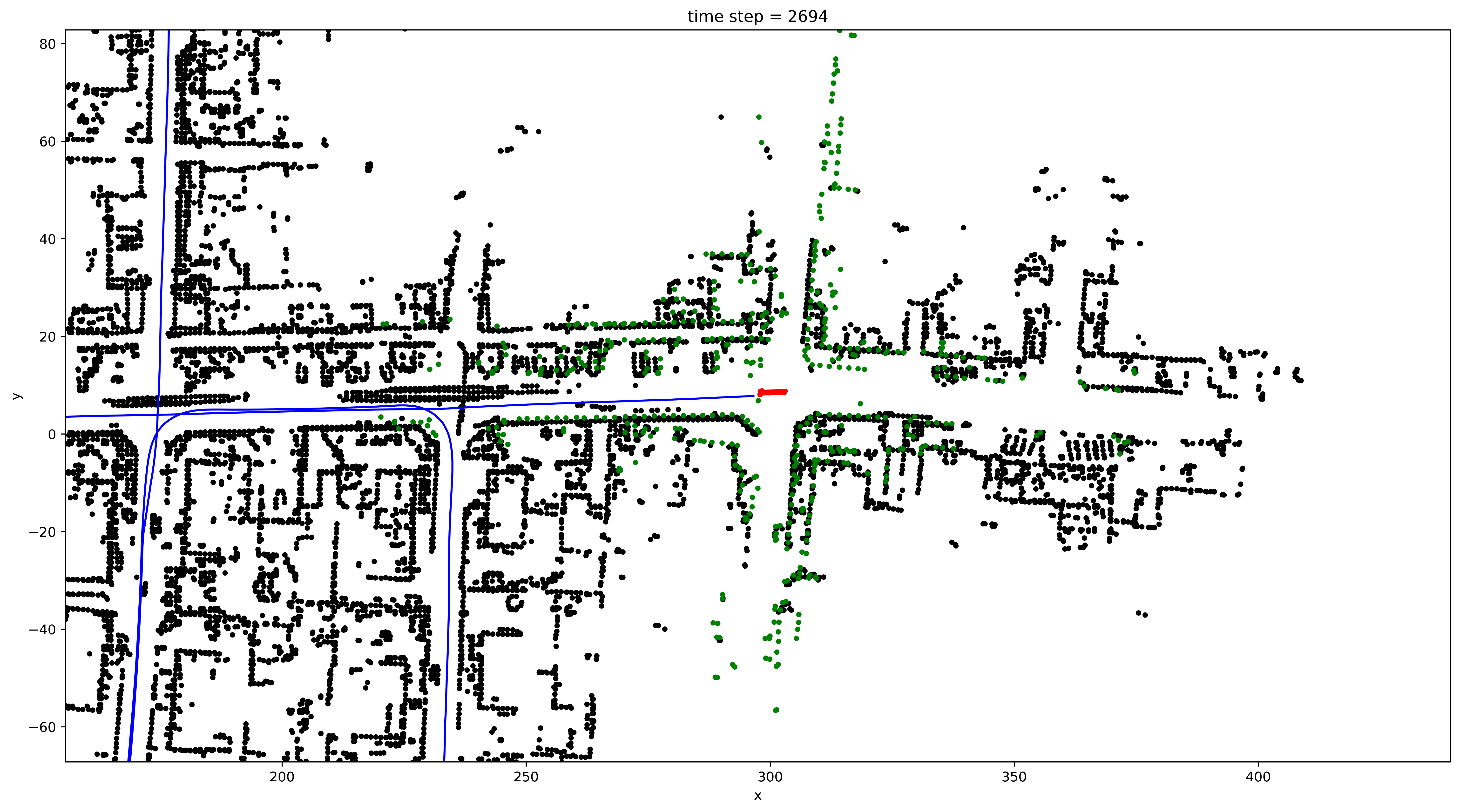}
			\label{snapshoteg1_9}}

\caption{Snapshots of a simulation at $k=2694$  }
		\label{snapshots1d}
\end{figure}

Figure \ref{snapshots1c} shows the snapshot at time step $k=670$, where the S2M threads happens to return a result. The matched and propagated result to time step $k=670$ clearly failed as seen in Figure \ref{snapshoteg1_7}. The S2M matching process can fail occasionally in very dense regions or regions of the map that have similar looking streets. This match is not accepted to update the pose of the particle, as the corresponding likelihood will significantly lower the weight of the particle compared to its original weight.  This simple check can quickly disregard failed matches. Figure \ref{snapshots1d} shows the snapshot of the simulation toward the end of map at time step $k=2694$. At this instant, the S2M block, which began at time step $k=2563$, resulted in a perfect match of the LIDAR scan and significantly improves the particle filter.

The proposed approach for localization can be used to continuously track the vehicle. This is often unnecessary due to the computational complexity of the localization process and one can switch to conventional tracking filters once the vehicle has been sufficiently localized.   In this paper, a vehicle is considered to be \emph{localized} if the standard deviation in the $[x,y,z]$ coordinates of the particle distribution is less than $10m$ for (a) at least $10$ consecutive time steps (LIDAR scans), and (b) the vehicle has moved a relative distance of at least $10m$ or (c) has turned a relative angle of $30$ degrees. Criteria (a) is useful to avoid spurious false positives, where the particle filter localizes to the wrong location with similar looking LIDAR scans. Criteria (b) and (c) are further necessary to make sure that there is enough diversity in the sequential LIDAR scans. This helps in reducing false positives. If the standard deviation in the position coordinates reduces to less than $10m$ but fails the joint criteria  of $\left((a)\: and\: ((b)\: or\: (c))\right)$, the vehicle has localized to a false positive region and the particle distribution is reset to the full map at the this time step. This way the localization procedure continues till the vehicle is localized, after which one can move to tracking mode. Note that, tracking is not considered in this paper as the main focus is on localization. The threshold of $10m$ is just chosen for simulation purposes and to indicate localization by bringing the uncertainty of the vehicle's position from $\pm300m$ to $\pm10m$.
  While stricter localization criteria can be considered ($<5m$), it is assumed that one can switch to tracking mode and thus avoid the computationally expensive S2M block within $\pm 5m$ of the true vehicle postion. Efficient filters, such as Extended Kalman filter (EKF) \cite{moore} or the Unscented Kalman Filter (UKF) \cite{jul1} can then be used for tracking as they  might be better suited from a computational standpoint than particle filters.  
  
\begin{figure}
		\centering
			\includegraphics[trim=4cm 1cm 1cm 1cm,clip=true,width=5.8in]{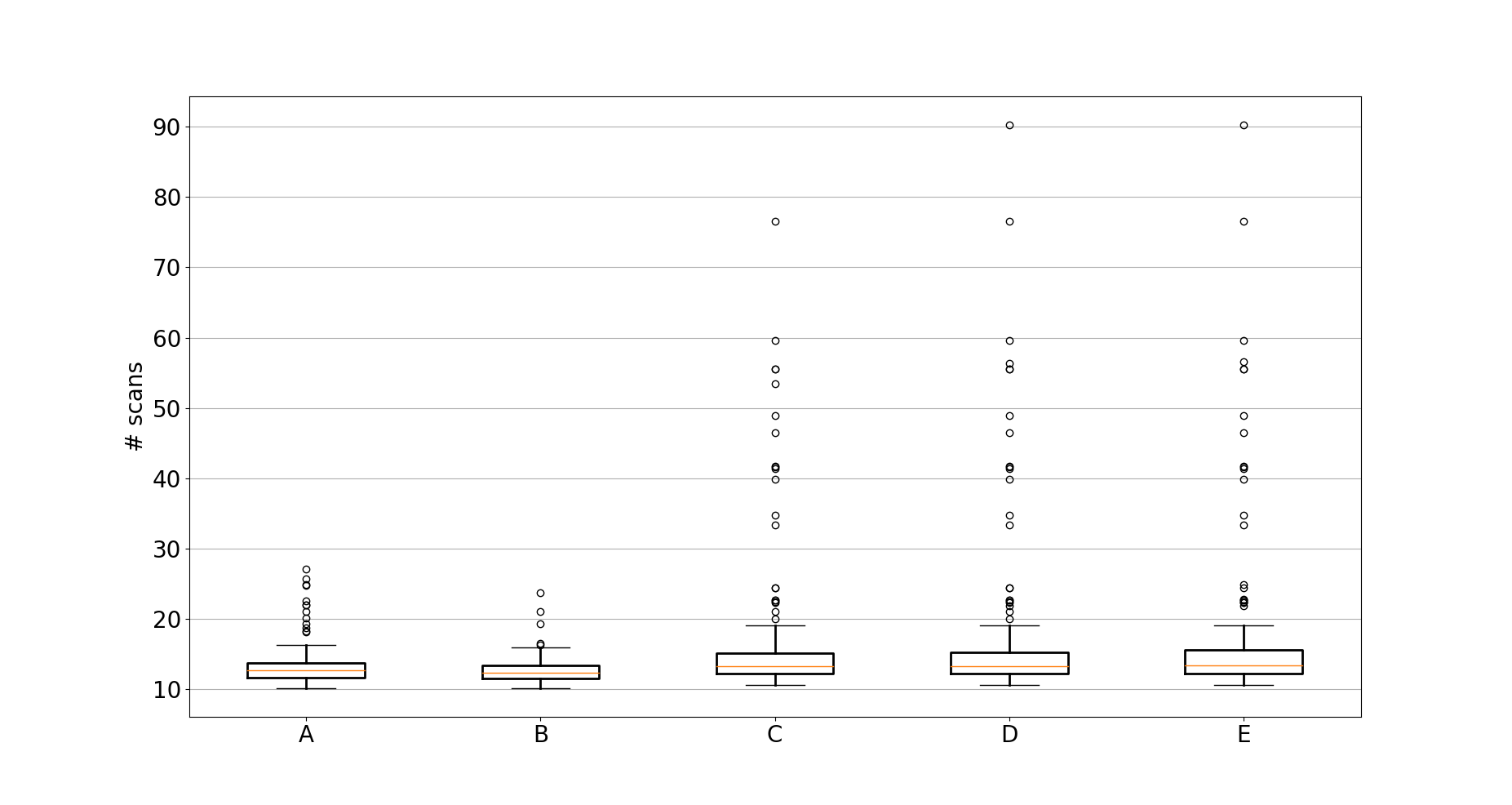} 

		\caption{Localization performance}\label{results11}
	\end{figure}

  Figure \ref{results11} shows the results of the simulation on sequence `05' of the KITTI odometry dataset. As the particles in the particle filter are randomly initialized, the simulations are repeated 250 times to generate statistical box plots of the localization performance. The localization performance is compared using five metrics (labeled as A,B,C,D and E in Figure \ref{results11}), with units as the number of sequential LIDAR scans (y-axis in Figure \ref{results11}) from time $t=0$. Here, the number of timesteps and number of LIDAR scans are used interchangeably as every timestep of the simulation corresponds to a LIDAR scan. There are a total of 2750 LIDAR scans in sequence `05' of the KITTI odometry dataset.  The boxplot (box and whiskers plot) in Figure \ref{results11}, labeled as Metric `B', shows the number of time steps (or LIDAR scans) required to first achieve localization, i.e. satisfy the localization criteria. It can be observed that the 75th percentile of the blox plot is around 14 scans and the largest outlier is about 24 scans. With 2750 total scans, the vehicle was localized within $1\%$ of the total scans. This indicates very good localization performance of the proposed approach. Once localized, tracking can be performed with more particles or other filters.
   
  In some cases, after the vehicle satisfies the localization criteria, the particle filter can occasionally diverge from the true vehicle state. This is because, the  1000 particles tend to coalesce around the particle with the highest weight. At this point, any sudden vehicle turn will become difficult to track  as there is lack of diversity among the particles. A straightforward remedy is to significantly increase the number of particles and/or increase process noise in the dynamical model if necessary. Alternatively, to establish the efficacy of the proposed localization approach, we detect loss in localization and reset the process at the same time instant.  The loss in localization is detected when the standard deviation of the position coordinates exceeds $50m$ (indicating a loss in track), even if the S2M block returns an exact match to the true the vehicle position. At this point, the particles of the particle filter are re-initialized over the whole map at this time step. This resetting approach will increase the time it takes or equivalently the number of scans for localization. Metric `A' in Figure \ref{results11}, captures this loss in localization over the complete 2750 time steps compared to Metric `B'. Metric `A' shows the number of scans it took to localize the vehicle according to the aforementioned localization criteria, including occasional resetting of the particle filter when it fails. Metric `A' is slightly above Metric `B', indicating that there were few instances where the particle filter was reset but eventually localized to the true vehicle position. The 75th percentile  is about 17, which shows that 75\% of cases were localized using only 15 scans (or timesteps), though some outlier cases took about 28 scans. Nevertheless, the approach was still able to localize the vehicle after particle filter divergence. The S2M block is able to provide a good match, that subsequently makes the particle filter distribution  converge to the true vehicle position. 
  
    Note that the localization criteria is chosen to emulate real-time conditions and does not use the true ground pose data from the KITTI dataset. The localization criteria only looks at the standard deviation of the vehicle's position  estimated by particle filter. An ideal offline performance metric should check if the localized vehicle's position is in fact close to the true vehicle position. To this end, the metrics `C', `D' and `E' in Figure \ref{results11} use ground truth data to measure localization performance. In addition to the localization criteria of $\left((a)\: and\: ((b)\: or\: (c))\right)$, metric `C' represents the number of scans before the mean position of the vehicle,  estimated by the particle filter, is  within $\pm 5m$ of the vehicle's true position for 10 consecutive timesteps. This is evaluated by finding the time step $k$ for which  $|\mathbf{p}_{est}(t)-\mathbf{p}_{truth}(t)|\le 5m$ for all $t_k \le t\le t_{k+10}$. The 75th percentile in the box and whiskers plot for metric `C' shows that $75\%$ of the cases were localized to their true vehicle positions for $10$ consecutive timesteps within $16$ timesteps. There are some outliers that took 78 timesteps (also counted as the number of scans) to reach true localization. It can be observed that there is a wide outlier disparity between metric `C' and metrics `A' or `B'. This is because the localization criteria in metric `A' uses a threshold of $10m$ for the standard deviation estimated by the particle filter, while metric `C' uses a stricter criteria of $\pm 5m$ about the true position of the vehicle. Even though the standard deviation of the particle filter is reduced, there is a small bias or lag in the mean compared to the true position. This is due to the S2M block which matches the pose of the LIDAR scan at a previous timestep and propagates it to the current time using the vehicles relative pose estimates. Once the S2M block corrects the particle with highest weight, subsequent resampling stages of the particle filter adds more particles around this particle. As more particles are pushed closer to the true pose of the vehicle, this bias in the mean estimate is eventually reduced, and thus some outlier cases take about 78 scans to get to the $\pm5m$ error about the true position of the vehicle. 
    
    Metric `D' is similar to metric `C', but it uses 50 consecutive steps to verify that the localization by the particle filter is in fact stable, or in other words the time step $k$ is determined for which  $|\mathbf{p}_{est}(t)-\mathbf{p}_{truth}(t)|\le 5m$ for all $t_k \le t\le t_{k+50}$. It can be observed from   Figure \ref{results11} that both metrics `C' and `D' are similar, implying that the localization to the true position of the vehicle ($\pm5m$ error) was achieved and even remained stable for 50 consecutive timesteps. Though the worst case took about 90 scans to reach an error of $\pm5m$.  After which, the S2M block can be switched off or other vehicle tracking algorithms can be used efficient. Finally, metric `E', which is similar to metric `D', has a stricter threshold of $2.5m$ for the error in the mean estimate. The time step $k$ for metric `E' is found when the condition  $|\mathbf{p}_{est}(t)-\mathbf{p}_{truth}(t)|\le 2.5m$ for all $t_k \le t\le t_{k+50}$ is satisfied. Metrics `D' and `E' look similar, except for the 75th percentile of metric `E' being slightly higher than the 75th percentile of metric `D'. This implies that the proposed localization approach is cable of localizing the vehicle to within $\pm2.5m$ of the true vehicle's position. 

		\subsection*{Brief note on implementation}
\begin{figure}
		\begin{center}
			\includegraphics[trim=4cm 1.8cm 1cm 1cm,clip=true,width=5.8in]{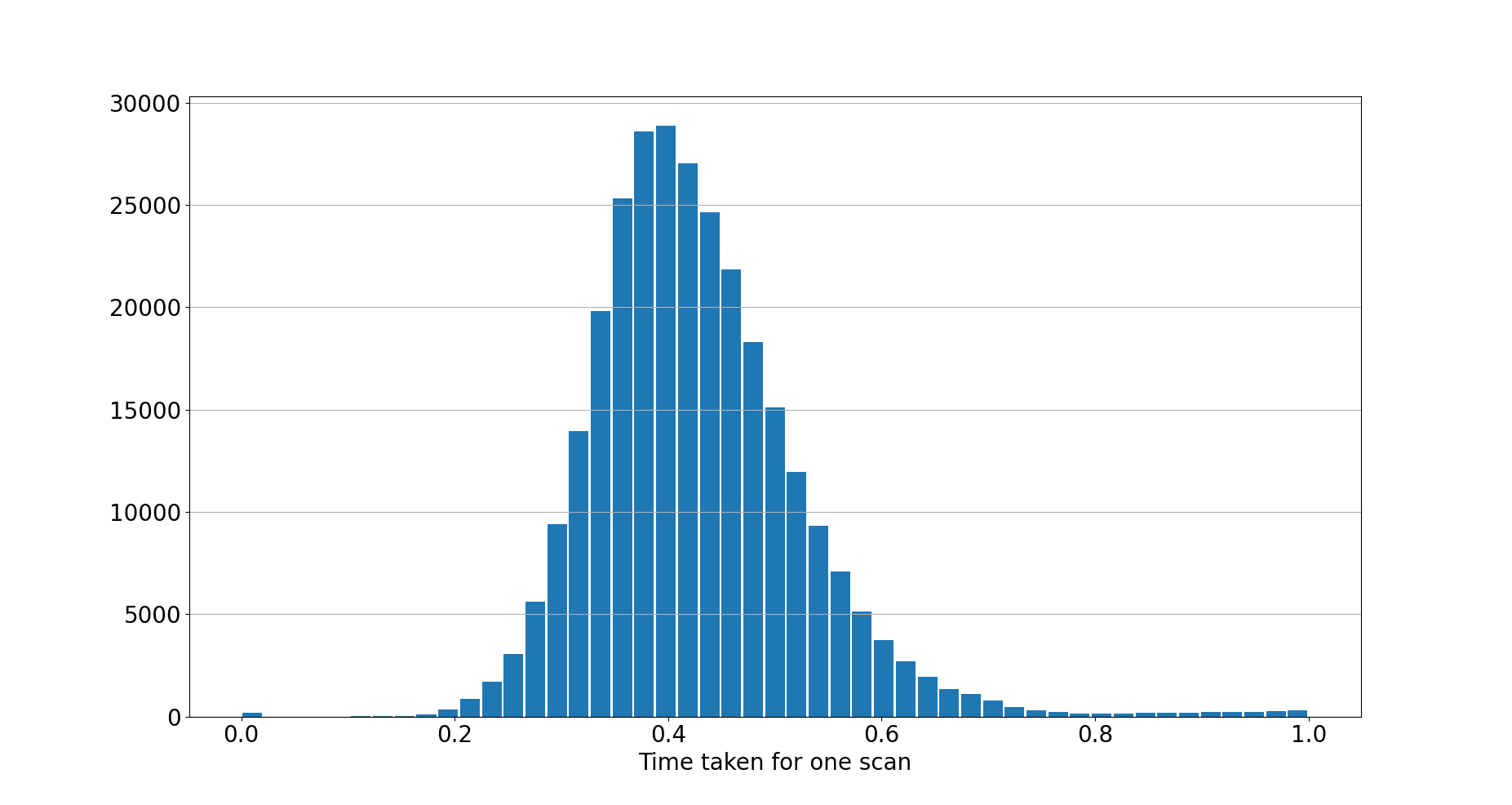}
		\end{center}
		\caption{Latency performance for current implementation (x-axis time in seconds) }\label{latencyhist11}
	\end{figure}

	As described in the Figure \ref{flchart}, the blocks run in parallel threads to provide fast particle filter estimates coupled with slower but accurate global pose estimates of the S2M block.  In the simulations an Intel i9 processor with 8 cores was used to simulate the proposed approach with 32GB of RAM. Point Cloud Library (PCL)  based implementation was used for the GICP in the RPE block and the second stage of the S2M block. The global nearest point to map distances $\mathcal{L}$ is implemented using parallel hash map \cite{parahash} for fast retrieval. Alternatively, sparse tensors can be used to save and retrieve the distances. The simulation reads the scans from the KITTI database in a  sequential loop, and the wall-clock time is recorded between scans. This  process is repeated over multiple runs using different initial conditions, and the recorded time between scans is shown using the histogram in Figure \ref{latencyhist11}. As the threads run in parallel, the time also includes the time taken by the  asynchronous S2M block.  It can be observed that the average time to process a scan is 0.4 seconds, with a Gaussian like distribution about the mean. It was observed that running the computationally intensive parallel routines on shared resources (CPU, RAM and caches) tends to make the overall process inefficient. It is to be noted that the objective of this paper is primarily to illustrate the validity of the proposed approach and not to establish high-performance implementation. Nevertheless,  a distributed architecture usign ROS2 \cite{abm6074} framework, where the computationally intensive scan-to-map matching routine is run on a separate machine will significantly improve frame-per-second (FPS). Further, GPU implementation can significantly improve the scan-to-map matching block and the GICP computations.  

	\comments{
	\begin{figure}
		\begin{center}
			\subfigure[Tracking steps]{\includegraphics[width=5in]{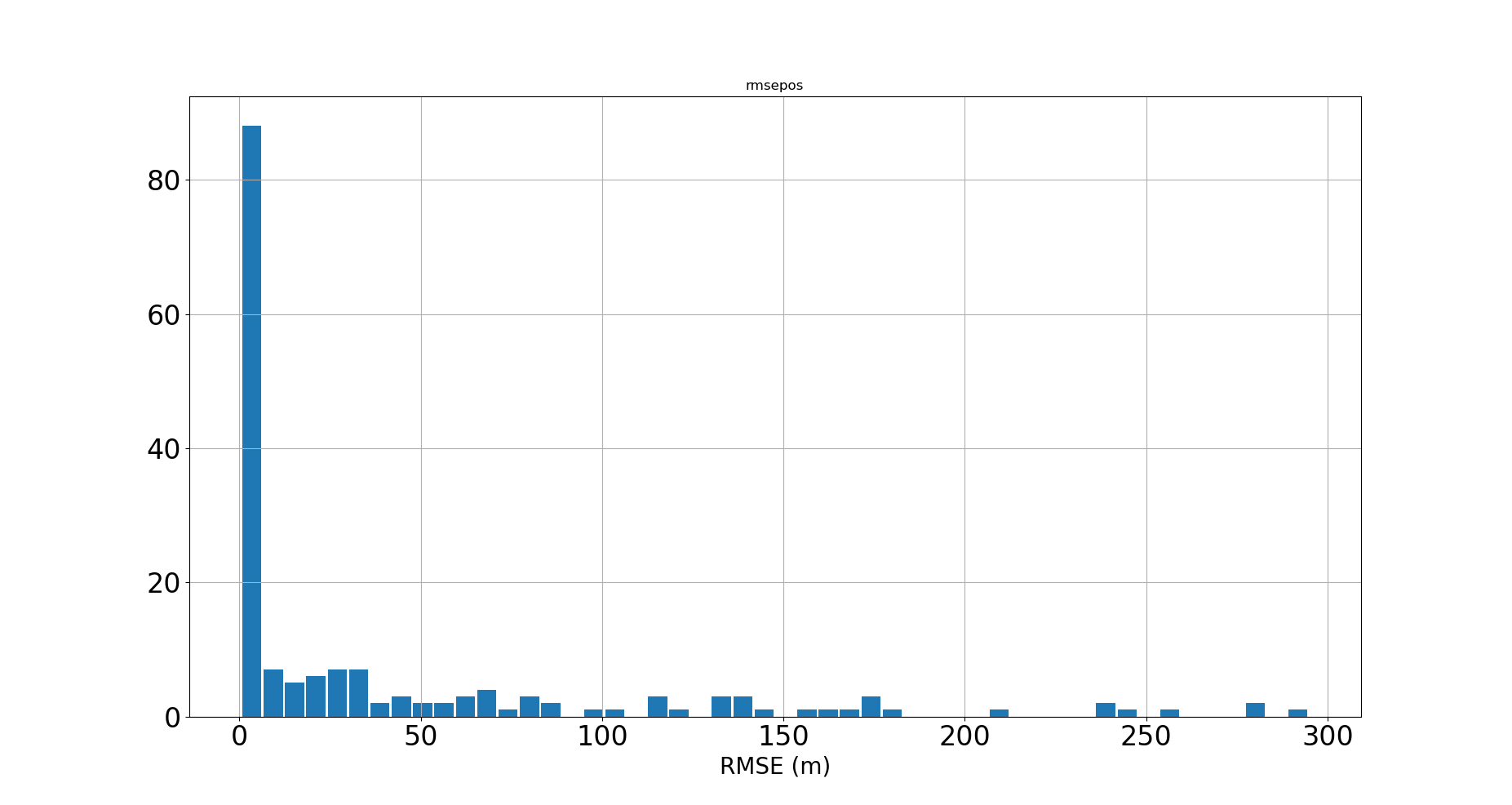}}
			\subfigure[Tracking steps]{\includegraphics[width=5in]{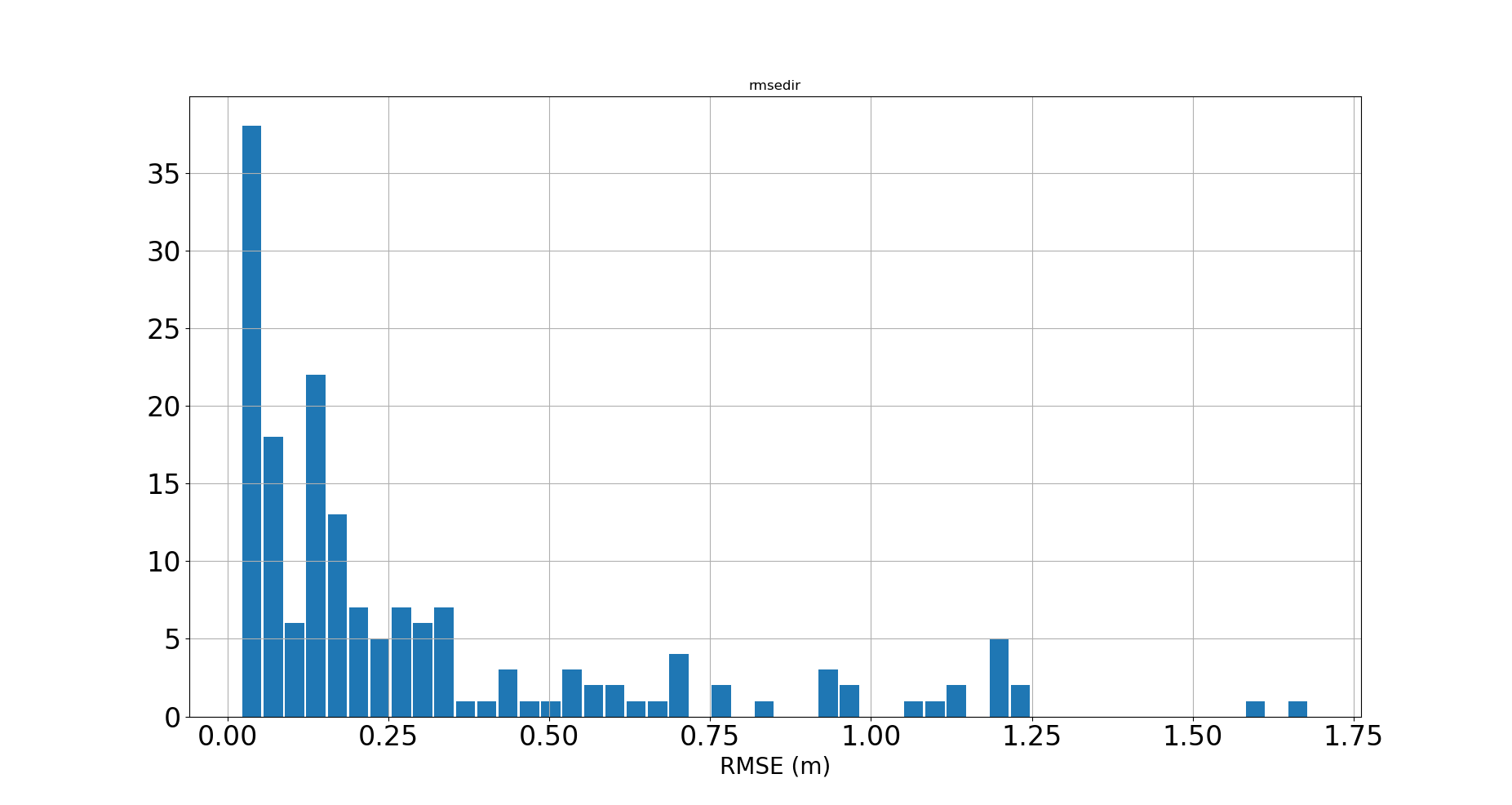}}
		\end{center}
	\caption{Tracking performance}
	\end{figure}
}

	\section{Discussion}
	
The results show that the scan-to-map matching process is highly advantageous, even when the corresponding computational complexity is high. The localization approach proposed in this paper, attempts to utilize the delayed result of the scan-to-map match by propagating the matched pose to the most recent  time instant. Due to errors in relative pose estimation, the propagated match may not be close enough to be used directly for the pose of the vehicle. Hence, the matched pose is only used to update the pose in the particles of the particle filter, which subsequently estimates the vehicle's full state. This symbiosis of the particle filter and the scan-to-map matching process can robustly localize the vehicle. This eliminates the need of excessively large number of particles that would often be required in  $3\Dim$. The approximate framerate achieved is only 2.5Hz (1/0.4s) which might be low for some applications. One approach is to decimate the LIDAR scan to few 100s or 1000s of points to improve the likelihood computational times. But we observed that, in dense maps like the KITTI dataset, decimation in fact degraded performance and increased the failure of the particle filter. This is because the decimated LIDAR point cloud for dense maps tend to have large false positive neighrest neighbors.  In our future work, we will implement GPU version of scan-to-map matching running on a separate machines with dedicated RAM and computing cores using the ROS2 framework.

	\section{Conclusion}
	The problem of  LIDAR-only localization and tracking approach was addressed by a hybrid approach of particle filtering and scan-to-map matching.  It was observed that with fewer particles, the approach was able to localize itself within few time steps (LIDAR scans). Further, the intermittent loss in tracking by the particle filter was corrected by the matching block leading to robust tracking performance. While the efficacy of the proposed approach was illustrated  on well structured KITTI odometry datasets, it is anticipated that these results provide strong optimism in developing high-performance GPU implementations that can be used for general urban areas and unstructured regions. Future work will assess the inclusion of IMU and vision sensors to further accelerate the localization process.

\end{document}